\documentclass[acmtog]{acmart}

\acmSubmissionID{331}

\setcopyright{acmcopyright}
\acmJournal{TOG}
\acmYear{2024}
\acmVolume{43}
\acmNumber{4}
\acmArticle{108}
\acmMonth{7} 

\usepackage{color}
\usepackage{xcolor}
\usepackage{graphicx}
\usepackage{booktabs}
\usepackage{amsmath, amsthm, amsfonts, amssymb}
\usepackage{mathrsfs}
\usepackage{textcomp}
\usepackage{epstopdf}
\usepackage{multirow}
\usepackage{wrapfig}
\usepackage{booktabs} %
\usepackage[linesnumbered,ruled,vlined]{algorithm2e}

\usepackage{ragged2e}
\usepackage[normalem]{ulem}
\usepackage{cleveref}
\usepackage{bm}
\usepackage{caption}
\usepackage{subcaption}
\usepackage{ulem}
\usepackage{enumitem}

\usepackage{booktabs, multirow} %
\usepackage{soul}%
\usepackage{xcolor} 
\usepackage{overpic}
\usepackage{tabularx}
\usepackage{makecell}

\newcommand{\eg}{\textit{e.g.}}

\definecolor{green}{rgb}{0, 0.5, 0}
\definecolor{orange}{rgb}{0.6, 0.3, 0.1}
\definecolor{red}{rgb}{1.0, 0.0, 0.0}
\definecolor{teal}{rgb}{0.0, 0.4, 0.4}
\definecolor{purple}{rgb}{0.65,0,0.65}
\definecolor{saffron}{rgb}{0.95,0.75,0.2}
\definecolor{turquoise}{rgb}{0.0,0.4,0.8}
\definecolor{brown}{rgb}{0.5, 0.16, 0.16}
\definecolor{brickred}{rgb}{.6, .2 .1}
\definecolor{coral}{rgb}{1,0.45,0.33}
\definecolor{newcolor}{rgb}{.8,.349,.1}

\let\maketitlesup\maketitle
\usepackage{xpatch}
\xpatchcmd{\maketitlesup}{\@mkteasers}{}{}{}
\xpatchcmd{\maketitlesup}{\@mkabstract}{}{}{}

\begin{document}
\title{Split-and-Fit: Learning B-Reps via Structure-Aware Voronoi Partitioning}

\author{Yilin Liu}
\email{whatsevenlyl@gmail.com}
\affiliation{%
	\institution{Shenzhen University}
	\country{China}	
}	
\affiliation{
	\institution{Simon Fraser University}
	\country{Canada}	
}
\author{Jiale Chen}
\email{chenjiale0303@gmail.com}
\affiliation{%
	\institution{Shenzhen University}
	\country{China}	
}
\author{Shanshan Pan}
\email{psshappystar@gmail.com }
\affiliation{%
	\institution{Shenzhen University}
	\country{China}	
}
\author{Daniel Cohen-Or}
\email{cohenor@gmail.com}
\affiliation{%
	\institution{Tel Aviv University}
	\country{Israel}	
}
\author{Hao Zhang}
\email{hao.r.zhang@gmail.com}
\affiliation{%
	\institution{Simon Fraser University}
	\country{Canada}	
}
\affiliation{%
	\institution{Amazon}
	\country{Canada}	
}
\author{Hui Huang}
\email{hhzhiyan@gmail.com}
\authornote{Corresponding author: Hui Huang (hhzhiyan@gmail.com)}
\affiliation{%
	\department{College of Computer Science \& Software Engineering}
	\institution{Shenzhen University}
	\country{China}	
}

\renewcommand\shortauthors{Y. Liu, J. Chen, S. Pan, D. Cohen-Or, H. Zhang, and H. Huang}

\begin{abstract}

We introduce a novel method for acquiring boundary representations (B-Reps) of 3D CAD models which involves a two-step process: it first applies a {\em spatial partitioning\/}, referred to as the ``split'', followed by a ``fit'' operation to derive a single primitive within each partition. Specifically, our partitioning aims to produce the classical {\em Voronoi diagram\/} of the set of ground-truth (GT) B-Rep primitives. In contrast to prior B-Rep constructions which were bottom-up, either via direct primitive fitting or point clustering, our Split-and-Fit approach is {\em top-down\/} and {\em structure-aware\/}, since a Voronoi partition explicitly reveals both the number of and the connections between the primitives. We design a neural network to predict the Voronoi diagram from an input point cloud or distance field via a binary classification. We show that our network, coined NVD-Net for neural Voronoi diagrams, can effectively learn Voronoi partitions for CAD models from training data and exhibits superior generalization capabilities. Extensive experiments and evaluation demonstrate that the resulting B-Reps, consisting of parametric surfaces, curves, and vertices, are more plausible than those obtained by existing alternatives, with significant improvements in reconstruction quality. Code will be released on \url{https://github.com/yilinliu77/NVDNet}.

\end{abstract}



\begin{teaserfigure}
  \centering
  \includegraphics[width=\linewidth]{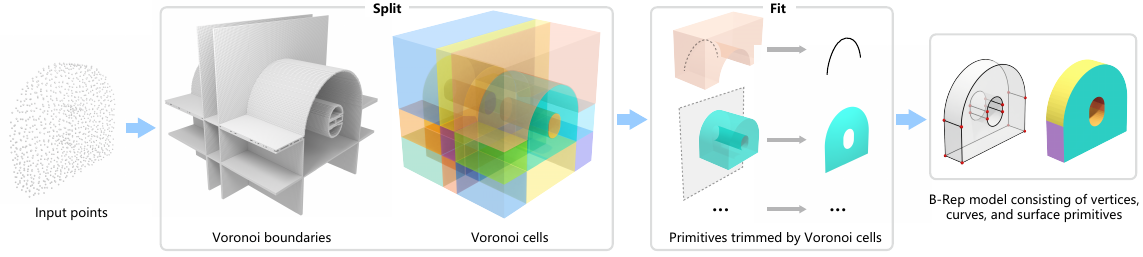}
  \caption{
     \textbf{Split}-and-\textbf{fit} offers a completely new perspective for acquiring B-Reps for 3D CAD shapes, e.g., from a point cloud. We first obtain a {\em spatial partitioning\/} of the volumetric space (i.e., the ``split'') and then fit a {\em single\/} primitive per partition.
     This is realized with the classical Voronoi diagrams.
  }
  \label{fig:teaser}
\end{teaserfigure}

\maketitle

\section{Introduction}
\label{sec:intro}

Computer-aided design (CAD) models play critical roles in design, engineering, manufacturing, and robotics applications.
The de facto and preferred shape representation for general 3D CAD models is the {\em boundary representation\/} or B-Rep~\cite{solidgen23,brepnet23,fayolle2023survey}.
A B-Rep describes a 3D solid by explicitly defining the limits of its volume in a structured and compact way through parametric surfaces, curves, vertices, and their 
topological relations. 
The wide use of B-Reps for CAD modeling and editing has generated much interest in B-Rep reconstruction from unstructured inputs such as 
point clouds or distance fields.

\begin{figure*}
    \centering
    \begin{overpic}[width=0.99\linewidth]{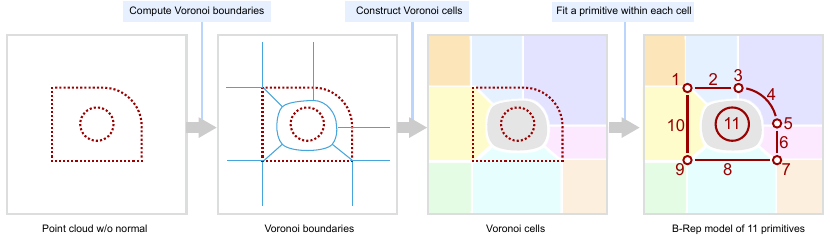}
    \end{overpic}
    \caption{
        Overview of our method. 
        Given an input model (i.e., a point cloud, mesh, or distance field), 
        we first compute the Voronoi boundaries of the underlying shape and subsequently construct the Voronoi cells. 
        Then we fit an elementary primitive for each Voronoi cell individually, with the cell boundary serving to naturally trim the primitive, and extract their corresponding connectivity from the Voronoi diagram. 
        Finally, we reconstruct the CAD model (5 vertices, 4 lines and 2 circles in the figure) in B-Rep by combining the primitives and their topological relations.
    }
    \label{fig:overview}
\end{figure*}

Classical approaches to B-Rep modeling over point clouds typically resort to clustering or primitive fitting via RANSAC~\cite{ransac,efficient_ransac,globfit}, region
growing~\cite{region_growing}, or variational shape approximation~\cite{variational04,variational12,variational20}.
Recent deep learning-based methods employ deep features for instance segmentation~\cite{parsenet20,hpnet21,primitivenet,sed23} or direct primitive 
detection~\cite{complexgen22}. All of these approaches are bottom-up and based on {\em local\/} geometric or topological features without explicit
optimization or supervision with respect to global structural properties such as primitive counts.
Unsupervised learning of constructive solid geometry (CSG) trees~\cite{kania2020ucsg,ren2021csgstump,capri22,d2csg23} has found some recent success for CAD 
modeling. However, all of these methods are trained to minimize the reconstruction error, %
which is not strongly tied to CSG tree optimization. Technically, there are infinitely many CSG trees which would yield zero reconstruction error. This causes an 
inherent {\em ambiguity\/} in the optimization setup. As a result, the obtained CSG constructions often contain unnatural and an excess of redundant primitives.

In this paper, we introduce a new perspective on acquiring B-reps for 3D CAD models which represents a significant departure from conventional approaches. Instead of
directly operating on an input cloud, either via clustering, segmentation, or primitive fitting/detection, we first perform a {\em spatial partitioning\/} of the volumetric space in 
a ``split'' operation. This is followed by a ``fit'' step to derive a {\em single\/} primitive within each partition to reproduce a primitive in the {\em ground-truth (GT)
B-Rep representation\/} of the input point cloud.
Specifically, we enforce our spatial partitioning to attain a {\em unique\/} and well-defined target, which is the classical {\em Voronoi diagram\/} of the set of GT B-Rep primitives. 
In contrast to prior bottom-up B-Rep constructions, our Split-and-Fit approach is {\em top-down\/} and {\em structure-aware\/}, since a Voronoi 
partition explicitly reveals both the number of and the connections between primitives.

We design a neural network to infer the Voronoi diagram from an input point cloud or distance field by training a binary classifier, which predicts whether
each voxel in 3D space lies on the boundary of a Voronoi cell or not based on local features. We show that our network, coined NVD-Net for neural Voronoi diagrams, 
can effectively learn Voronoi partitions for CAD models from training data. By confining the per-cell single primitive fitting to Voronoi cell boundaries, the primitives are
automatically trimmed without needing a separate, intricate process as in other works such as ComplexGen~\cite{complexgen22}.
Also importantly, our learning problem does not suffer from the ambiguity issues for neural CSG constructions, since our training target, the GT Voronoi diagram,
is unique.

Our contributions can be summarized as follows:
\begin{itemize}
\item Introducing Split-and-Fit, a novel paradigm for B-Rep reconstruction through spatial partitioning of volumetric space, which is top-down and structure-aware.
\item NVD-Net, a deep neural network for neural Voronoi diagram prediction from point clouds or distance fields.
\item An efficient scheme to extract B-Rep surfaces, curves, vertices, and their connectivities from a Voronoi diagram. 
\end{itemize}

We train our NVD-Net and test our B-Rep reconstruction method on the ABC dataset~\cite{ABC}, which offers GT B-Reps.
Extensive experiments and evaluation demonstrate that compared to state-of-the-art alternatives, our method produces more plausible B-Rep constructions with
lower geometric errors, higher topological consistency, and improved instance identification, while exhibiting superior generalization capabilities over unseen 3D shapes.

\section{Related Work}
\label{sec:rw}

The reconstruction of CAD models has been widely studied in the past decades.
Early attempts either use analytical or learning-based methods to fit individual primitives from the input point cloud.
Recently, some methods have been trained to predict intermediate representation to facilitate CAD reconstruction.
We summarized the three categories of methods in the following.

\paragraph{Shape fitting}
Traditional methods usually reconstruct CAD models by detecting the primitives (e.g. planes, cylinders, spheres, cones, torus, etc.) from input point clouds. 
They either use RANSAC~\cite{ransac,efficient_ransac} or Region Growing~\cite{region_growing, OesauLA16} to detect these primitives, which are then merged into a single CAD model by inferencing the topologies between them~\cite{poiyfit,ksr}.
Li et al.~\shortcite{globfit} proposed a method to further constrain the detected primitives with the global relation between primitives. 
Also, variational shape approximation~\cite{variational04,variational12,variational20} has been adopted to reconstruct primitives from point clouds.
The explicit optimization of the shape partitions enables better modelling of the shape relations.
Additionally, Attene et al.~\cite{AtteneP10} adopt a "fit-and-segment" pipeline, iteratively clustering triangle faces based on local primitive fitting.
This process enables the extraction of primitives from the underlying CAD models based on the generated clusters.
Further details can be found in the survey on basic shape primitive fitting~\cite{KaiserZB19}.
However, these methods individually address geometric and topology properties, making the reconstruction process easily stuck in a local minimum.

\paragraph{Point segmentation}
Recent learning-based methods~\cite{SPFN19, LiSDYG19,parsenet20,hpnet21,primitivenet,bpnet23,sed23} used a "segment-and-fit" pipeline to reconstruct CAD models from point clouds.
These methods train a point-based neural network to perform instance segmentation on the input point cloud, and then fit primitives to the segmented clusters.
Specifically, SPFN and ParSeNet~\cite{SPFN19,parsenet20} embed the input points into a representation space, where the feature code for points that belong to the same primitive are closer to each other.
After a mean-shift clustering, the clustered points are fed into the fitting module to produce the final geometry.
HPNet~\cite{hpnet21} and SEDNet~\cite{sed23} further enhance the representation learning by hybrid shape descriptors and multi-stage feature fusion mechanisms.
NerVE~\cite{nerve23} also proposed directly predicting the structured edges instead of the pure point segmentation.
Meanwhile, Point2CAD~\cite{point2cad} further employs an analytic-neural reconstruction method to fit and recover the structured CAD models based on the segmentation. 
However, all these methods have been designed to approximate the underlying shape in a bottom-up manner, which is also prone to local minima.
The mixed combinatorial assignment of point labels and continuous parameter fitting also makes the learning process unstable.

\paragraph{Direct CAD learning}
Instead of fitting primitives, some other representations have been adopted for CAD reconstruction.
Li et al.~\shortcite{secadnet23} and Lambourne et al.~\shortcite{lambourne22} attempted to learn 2D sketches and 3D extrusion parameters from point clouds, which is conventional in traditional CAD modelling.
DEF~\cite{DEF22} learns to construct a distance-to-feature field to represent the input range scan and then reconstruct feature curves in CAD models.
BSPNet~\cite{bspnet20} was trained to predict a set of planes to build a binary space partitioning tree, where the learning process is motivated by minimizing a reconstruction error.
The resulting planes can be assembled into a watertight mesh to represent the underlying shape.
CAPRINet~\cite{capri22} and D2CSG~\cite{d2csg23} further enhance the partitioning by supporting quadric primitives and performing single-object optimization to produce better CSG-Trees. 
Additionally, Xu et al.~\cite{zonegraph21} present a method to infer and reconstruct the modelling sequence of a CAD model by utilizing zone graphs to represent the spatial partitioning induced by the model's BRep faces.
However, while the reconstruction error is minimized, the structure of the CSG-Tree is often far from optimal.

Recent advances in autoregressive models and transformer architectures have paved the way for the direct generation of CAD models in Boundary Representations (B-Reps).
PolyGen~\cite{polygen20} employs a transformer-based pointer network to sequentially generate mesh vertices and faces, with its probabilistic model design facilitating the creation of novel structures from diverse inputs. 
Building upon this, SolidGen~\cite{solidgen23} extends the capabilities to B-Rep CAD models, incorporating elementary surfaces such as cylinders or cones. 
Utilizing the Indexed Boundary Representation framework, SolidGen methodically produces vertices, curves, faces, and subsequently converts them into B-Rep models. 
Furthermore, MeshGPT~\cite{meshgpt23} has introduced an autoregressive generation of compact meshes with sharp edges. 
Unlike direct prediction of surfaces and curves, MeshGPT's learning process operates in a pre-quantized latent space. 
However, while these methods succeed in generating compact meshes, they usually lack parametric surfaces and curves~\cite{polygen20,meshgpt23}, which are essential for subsequent editing and rendering tasks. 
Although SolidGen~\cite{solidgen23} outputs parametric surfaces, aligning the latent code of the condition with the generative model remains a challenge, often leading to outputs that are inconsistent with the input conditions and lacking in detail.

In contrast, ComplexGen~\cite{complexgen22} formulates CAD reconstruction as a detection task, directly determining the parameters and topologies of each primitive via a conventional object detection framework. 
This approach includes an optimization step to refine the topology and geometry of the inferred primitives. 
Nonetheless, their learning process is prone to local minima due to the indeterminate number of primitives, and the mixed topological and geometric optimization complicates the reconstruction process.

While \textit{Point-based} Voronoi diagrams are prevalent in computer graphics and computational geometry, with applications in mesh reconstruction~\cite{AlliezCTD07,voromesh23}, Medial Axis Transform (MAT) computation~\cite{wang22}, and mesh simplification~\cite{voronoi_simplify}, these diagrams have seen limited application in CAD model reconstruction. 
SEG-MAT~\cite{segmat} utilizes a form of Voronoi Diagrams, the Medial Axis Transform (MAT), for shape representation and segmentation. 
However, this approach is restricted to part segmentation and does not generate structured CAD models.
Our method leverages the \textit{primitive-based} Voronoi diagram as an intermediary representation in reconstructing B-Rep models. 
The construction of the Voronoi diagram, achieved through a simple binary classification of Voronoi boundaries, along with the straightforward inference of topological relations from the connectivity of Voronoi cells, significantly reduces ambiguity in the learning process.

\section{Preliminaries}
\label{sec:overview}

\begin{figure*}
    \centering
    \includegraphics[width=\linewidth]{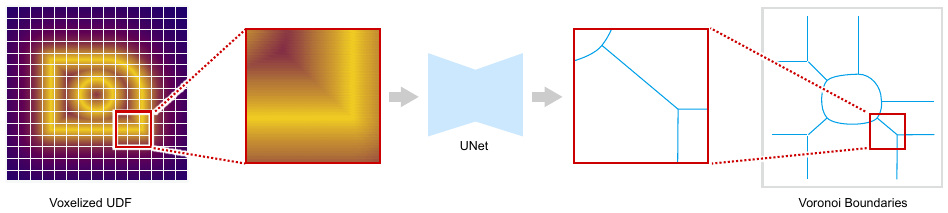}
    \caption{
        Overview of our network. 
        We first voxelize the input UDF field. 
        Each voxel contains 4 features $(d, g_x, g_y, g_z)$ which indicates the UDF value and the gradient vector of the UDF field.
        The voxel grid is split into overlapped local patches and fed into the standard UNet individually.
        A single channel binary flag is predicted within each voxel to indicate whether this voxel contains any Voronoi boundary.
    }
    \label{fig:network}
\end{figure*}

\paragraph{Definition of the Voronoi diagram over B-Reps.}
Unlike most previous methods~\cite{voromesh23,voronoi_simplify,voronoinet20, wang22, AlliezCTD07}, which define Voronoi diagrams on point sets, our Voronoi diagram is defined on primitives (vertices, curves and surfaces). 
Similar to SOTA methods~\cite{bspnet20,capri22,d2csg23,complexgen22}, our primitives are restricted to planes, spheres, cylinders, cones, torus as surfaces and lines, circles, ellipses as curves.
Given a set of primitives, the Voronoi diagram $G_v(N_v, E_v)$ is a partitioning of the volumetric space, which consists of a set of adjacent Voronoi cells $N_v$. 
Here, $E_v$ is the adjacent matrix that indicates whether two Voronoi cells are neighbouring.
Based on the bisection property of Voronoi diagrams, any point inside one cell should always have the closest distance to its corresponding primitive than to other primitives. 
Since the primitives and their corresponding Voronoi diagram are dual structures, we do not need to store explicit information about the primitives (\eg,~type, parameters, etc.).
Instead, we store the dual Voronoi diagram.
Storing the dual structure gives us the freedom to adapt to various types of primitives during learning.
More importantly, it allows us to convert a surface-based BRep model to a volumetric representation (described in Section~\ref{sec:method}).
Learning on a surface representation is vulnerable due to a lack of a suitable representation to simultaneously indicate a mixture of geometric, topological, and combinatorial features. 
In contrast, learning a volumetric representation through binary classification is more straightforward and robust.
Additionally, unlike the CSG-Tree used in previous CAD reconstruction methods~\cite{bspnet20,d2csg23,capri22}, the Voronoi diagram remains unique, reducing ambiguity in the training process.

\paragraph{Problem Statement and notions.}
The input to our method can either be a sparse 3D point cloud with or without normals, i.e., 
$P=\{p_i\}_{i=1}^k$, where $p_i \in \mathbb{R}^3 \text{ or } \mathbb{R}^6$,
a mesh, or a continuous distance field $f \colon \mathbf{X} \to \mathbb{R}$.
Our goal is to reconstruct the \textit{unique} Voronoi diagram of the GT primitives, which is then used to extract primitives and their connectivities for B-Rep model reconstruction (as shown in Fig.~\ref{fig:overview}). 
Notationally, we denote a B-Rep model as $M(V,E,F,\partial,\mathcal{P})$, where $V=\{v_i\}_{i=1}^k$ is a set of vertices, $E=\{e_i\}_{i=1}^k$ is a set of edges, and $F=\{f_i\}_{i=1}^k$ is a set of surface patches.
$\partial_n, n=1,2$, represents the topological relations between vertices, edges, and surface patches (e.g., $\partial_2 f_i \in E$ represents the boundary of surface patch $f_i$).
We additionally recover the connectivity between surface patches to further facilitate the reconstruction of curves~\cite{sed23}.
And $\mathcal{P}$ represents the geometric properties of each primitive.

\paragraph{Method Overview.}
To compute the Voronoi boundaries for a given point cloud,
we first convert the input point cloud into an unsigned voxelized distance function (UDF) field via Neural Dual Contouring~\cite{ndc22}. As shown in  Fig.~\ref{fig:network}, we train a UNet-like neural network to predict the Voronoi diagram from the UDF field.
In order to recover the Voronoi diagram, our network is trained to identify the Voronoi boundaries from the UDF field  (described Sec.~\ref{sec:method_voronoi}).
Based on the predicted Voronoi boundaries, we employ a region-growing strategy to extract the Voronoi cells and their connectivities.
Inside each Voronoi cell, we fit a primitive using an analytical method and then reconstruct the CAD models in B-Rep by combining the primitives and their topological relations, described in Sec.~\ref{sec:method_primitive}.

\section{Methodology}\label{sec:method}

\subsection{Voronoi Diagram Prediction}
\label{sec:method_voronoi}
Our Voronoi predictor takes a voxelized UDF field $f$ as input and predicts the Voronoi diagram $G_v(N_v, E_v)$.
We use UDF fields as input instead of previously used point clouds~\cite{parsenet20,complexgen22,sed23},
since a UDF is a continuous function defined over the entire volumetric space, which is better suited for our space partitioning problem.
Also, point clouds can be easily converted into UDF fields~\cite{ndc22}.

\begin{figure*}
    \centering

    \begin{subfigure}[t]{0.15\textwidth}
        \centering
        \includegraphics[width=\textwidth]{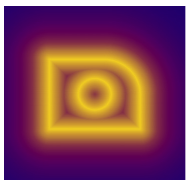}
        \caption{UDF field $f$}
    \end{subfigure}
    \hfill
    \begin{subfigure}[t]{0.15\textwidth}
        \centering
        \includegraphics[width=\textwidth]{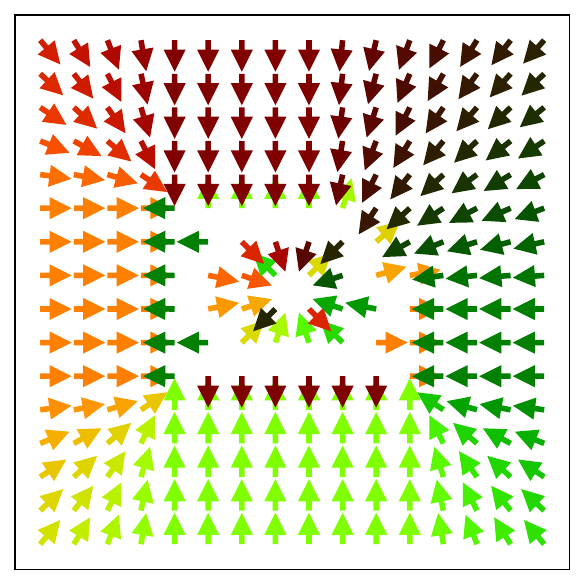}
        \caption{$f$'s 1st derivative $f^{'}$}
    \end{subfigure}
    \hfill
    \begin{subfigure}[t]{0.15\textwidth}
        \centering
        \includegraphics[width=\textwidth]{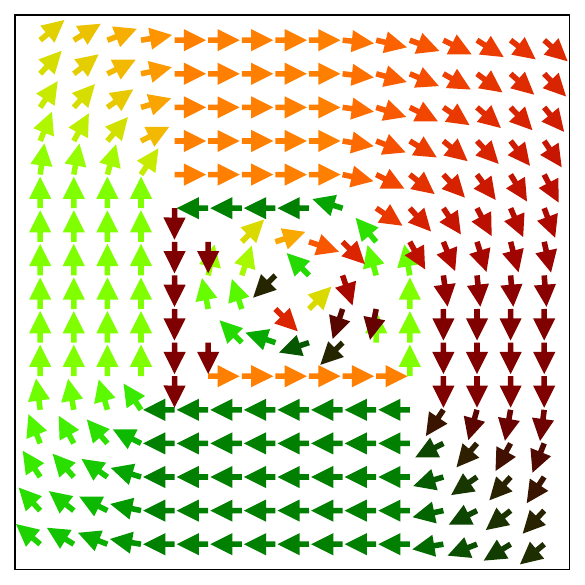}
        \caption{Orthogonal direction $f^{+}$ of $f^{'}$}
    \end{subfigure}
    \hfill
    \begin{subfigure}[t]{0.15\textwidth}
        \centering
        \includegraphics[width=\textwidth]{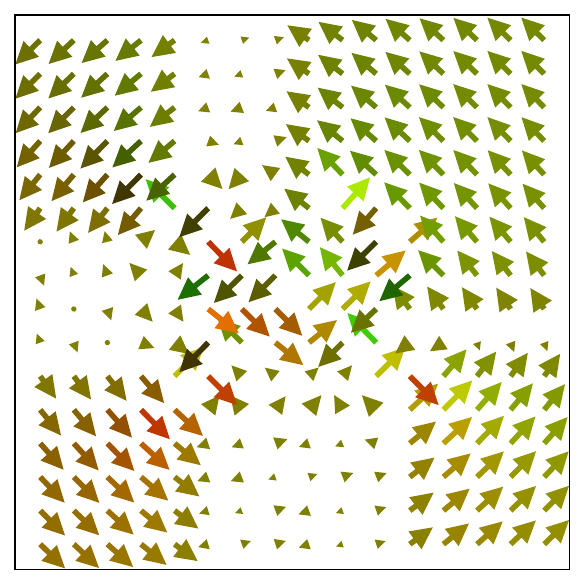}
        \caption{$f$'s 2nd derivative $f^{''}$ along $f^+$}
    \end{subfigure}
    \hfill
    \begin{subfigure}[t]{0.15\textwidth}
        \centering
        \includegraphics[width=\textwidth]{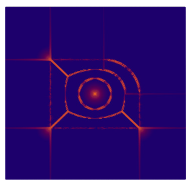}
        \caption{Norm of $f$'s 3rd derivative $|f^{'''}|$ along $f^+$}
    \end{subfigure}
    \hfill
    \begin{subfigure}[t]{0.15\textwidth}
        \centering
        \includegraphics[width=\textwidth]{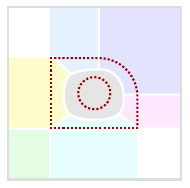}
        \caption{Voronoi cells $N_v$}
    \end{subfigure}

    \caption{
        The relation between the Voronoi boundaries and the derivative of the UDF field.
    }
    \label{fig:derivative}
\end{figure*}

Given a UDF field $f$, we want to predict a binary field $b \colon X \to \{0,1\}$, where $b(x)=1$ indicates that $x$ is on the Voronoi boundary, and $0$ otherwise.
In order to train such a network, we first discretize the UDF field $f$ into a voxelized grid $V_{r*r*r}$ with a fixed resolution $r$ ($r=256$ in all our experiments).
Each voxel contains a 4-channel feature code $(d, g_x, g_y, g_z)$, where $d$ is the UDF value and $g_x, g_y, g_z$ are the first-order derivatives of the UDF field, also denoted as the gradient vector.
Followed by a UNet-like network $F(V)$, the 4-channel feature code is mapped to a binary flag $b$ for each voxel, indicating whether it is on the Voronoi boundary; see Fig.~\ref{fig:network}. 
The detailed structure of the network is shown in the supplementary material.

Inspired by Neural Dual Contouring~\cite{ndc22}, we only leverage local features to predict Voronoi boundaries.
We split the whole voxel grid into $N$ local patches with a fixed stride $s$ and size $k$ ($s=16$ and $k=32$ in all our experiments).
Note that the local patches are overlapped, where each voxel can be contained in multiple such patches.
Patch overlapping can ensure that the prediction on the boundary of the local patches is consistent.
The network is trained to minimize the focal loss~\cite{focal20} between the predicted flag $b$ and the ground truth flag $b^*$,
\begin{equation}
    \mathcal{L}_{focal} = -\alpha (1-b^*)^\gamma \log(1-b) - (1-\alpha) b^* \log(b),
\end{equation}
where $\alpha$ and $\gamma$ are the hyper-parameters of the focal loss.
Please refer to the supplementary material for more details about the data preparation and training process.

\paragraph{Generalizability}
It is worth noting that ComplexGen~\cite{complexgen22} leverages a triple-branch transformer architecture to predict the parameters and topologies of each primitive.
Different cross-attention modules ensure full information exchange between the voxel and primitive features.
However, the comprehensive information exchange also makes the learning process hard to optimize, which might lead to poor generalizability; see Sec.~\ref{sec:generalization} for further details and a comparison.
In contrast, our prediction of Voronoi diagrams only relies on local geometric cues, which can significantly improve the model's generalizability.

\paragraph{Discontinuity over UDF}
Estimating Voronoi boundaries is similar to identifying discontinuities over the UDF field, which only relies on local geometric features; see Fig.~\ref{fig:derivative}.
Overall, the distance field of a CAD model is a piece-wise smooth function.
The discontinuity of the distance field is usually related to a change of the nearest primitive.
To find the discontinuity, we first calculate the derivative of the UDF field.
Fig~\ref{fig:derivative}(a) and Fig~\ref{fig:derivative}(b) show the original UDF field and the first-order derivative.
We can see that for points whose nearest primitive is a plane, their gradient vectors are parallel.
Thus, the discontinuity is always located at the joint of two planes.

We further calculate the second- and third-order derivatives of the UDF field to identify the joint between two quadric primitives.
Since the first-order derivative of the UDF field is a vector field, we choose the orthogonal direction of each gradient vector and calculate the second- and third-order derivatives along this direction in Fig~\ref{fig:derivative}(d).

We visualize the $L_2$-norm of the 3rd derivative in Fig~\ref{fig:derivative}(e), and we can see the discontinuity with high $L_2$-norm is located at the joint of two quadric primitives, which is consistent with the Voronoi boundaries of the target shape.
Thus, identifying the Voronoi boundaries is equal to finding the discontinuity of the second derivative of the UDF field, as shown in Fig~\ref{fig:derivative}(f).
However, real-world UDF fields are usually noisy, which makes the discontinuity harder to identify.
Thus, we choose to predict the Voronoi boundaries using a neural network to approximate this process, leading to improved robustness against minor local disturbances in the input data.

\subsection{Extraction of primitives and topologies}
\label{sec:method_primitive}
Based on the predicted Voronoi boundaries, we first conduct region growing in the volumetric space to reconstruct the Voronoi cells $N_v$.
We select a seed voxel $v_s$ and grow the region by adding the neighbour voxels $v_n$ with flag $b(v_n)=0$.
The region growing process is repeated until all the neighbour voxels are added.
The voxel grids will be clustered into regions, denoted as the Voronoi cells $N_v$.
The connectivity of the Voronoi cells $E_v$ can be easily inferred from neighbourhood relations of $N_v$.

By definition of the Voronoi diagrams, each Voronoi cell contains exactly one primitive.
Based on the UDF field, we use the least-square method~\cite{GTE} to fit a primitive inside each Voronoi cell.
For each Voronoi cell, we iterate over all the possible primitive types and choose the one with the lowest fitting error.
Note that the fitting process is only conducted in each Voronoi cell \textit{individually}.
Unlike ComplexGen~\cite{complexgen22} and search-based methods~\cite{globfit,ksr}, we do not need to consider assigning each point to a specific primitive.
Thus, the fitting process is considerably less ambiguous and more robust.

Since the connectivity of the Voronoi cells $E_v$ is already known, we can easily infer the topological relations $\partial$ between primitives.
For each primitive $i$, we find its adjacent Voronoi cells $N_{i}$ and examine whether the distance between the point inliers of the two primitives is less than a prescribed threshold. 

After extracting the primitives and their topologies, we can reconstruct the CAD models in B-Rep by combining the primitives and their topological relations.
Similar to SEDNet~\cite{sed23}, we additionally calculate the intersection curve between two adjacent surfaces to add missing curves during the fitting process.
The resulting B-Rep model can be edited, meshed, and visualized by most CAD software.
Please refer to the supplementary material for more details about the detailed explanation and the pseudocode of fitting.

\section{Experimental Results}\label{sec:results}

\subsection{Dataset and metrics}

We benchmark our method against leading CAD model reconstruction techniques on the ABC dataset~\cite{ABC}. 
Our comparative analysis includes state-of-the-art methods such as ComplexGen~\cite{complexgen22}, HPNet~\cite{hpnet21}, SED-Net~\cite{sed23}, and Point2CAD~\cite{point2cad}, as well as the traditional fitting method RANSAC~\cite{ransac,efficient_ransac}. 
Our model was trained on a dataset identical to that used by ComplexGen, HPNet, and SED-Net, comprising approximately 20,000 models featuring elementary and B-Spline primitives.

For a fair and comprehensive assessment, all methods were tested on the same test set of around 1,000 models. 
This test set included models with surface types such as Plane, Cylinder, Sphere, Cone, and Torus and curve types like Line, Circle, and Ellipse. 
Notably, B-Spline primitives were excluded from the test set due to observed instabilities in their analytical fitting from point data.

Nevertheless, it is essential to underscore that our primary contribution, the learning-based Voronoi Partition process, is agnostic to the actual primitive type. 
This is further elaborated in Section~\ref{sec:limitation}.

\paragraph{Evaluation Metrics:}
To assess each method's performance, we employ three metric categories:
\begin{itemize}
\item \textbf{Geometric Error:}
The Chamfer Distance (CD) and Light Field Distance (LFD)~\cite{lfd} are utilized for quantifying geometric inaccuracies and visual fidelity discrepancies between the reconstructed and ground truth CAD models. 
Notably, models with high accuracy but fragmented primitives can still display minimal geometric error. 
Hence, following ComplexGen~\cite{complexgen22}, detection metrics are also employed for a more nuanced performance evaluation.
\item \textbf{Number of Effective Primitives:}
As an additional metric, we report the count of effective primitives in the reconstructed CAD models, providing insight into the model's structural complexity.
\item \textbf{Detection Score:}
The averaged Precision, Recall, and F1-Score are used to assess the accuracy of primitive instances within the methods. 
This metric offers a complementary perspective to geometric error, facilitating a direct comparison between predicted and ground truth primitives. 
We apply multiple thresholds (0.1, 0.05, 0.02, 0.01, 0.005) for a thorough evaluation.
\item \textbf{Topological Error:}
To gauge the accuracy of predicted relationships among surfaces, curves, and vertices, we employ the F1-Score as a measure of topological error.
\end{itemize}

\subsection{Comparison}
We sample 10k points on the GT mesh as the input to all the methods.
All methods are tested with normal input except for our method.
In order to benchmark our approach against three baselines, we first extract the trimmed mesh for each reconstructed primitive. 
Then we sample points on the surfaces and curves and compute the previously mentioned metrics.
For RANSAC, we use the implementation from the CGAL library\footnote{\url{https://doc.cgal.org/latest/Shape_detection/index.html}}.
We report the results of both default parameters (\textit{RANSAC Default}) and the best parameters we tuned (\textit{RANSAC Tuned}), where $\epsilon=\{1\%*bbox\_diag, 0.001\}, prob=\{0.01,0.001\}, min\_points=\{1\%, 0.5\%\}, \epsilon_{normal}=\{0.9,0.9\}$, respectively.
For HPNet~\cite{hpnet21} and SEDNet~\cite{sed23}, we employ Point2CAD~\cite{point2cad} to extract the mesh from the point segmentation. 
As for ComplexGen~\cite{complexgen22}, we utilize its official implementation for extracting the trimmed models.
The points essential for metric computation are derived from the trimmed meshes using Poisson disk sampling. 
The quantity of these sample points is determined based on the area of the surfaces or the length of the curves.
Quantitative comparisons between each method are detailed in Table~\ref{table:geometric_error}, \ref{table:topological}, \ref{table:area_detection}, \ref{table:point_detection}.
For qualitative analysis, we have chosen 30 \textbf{representative} CAD shapes from our test set, showcasing a variety of geometric features such as standard furniture, high-genus structures, and complex compositions of planes, cylinders, cones, and open surfaces. 
A subset comprising five examples is presented in Fig.~\ref{fig:qualitative_representative}, with additional details available in the supplementary material.
To ensure an unbiased presentation, we have also \textbf{randomly} selected 30 cases from the test set for visualization, avoiding any manual cherry-picking.
Five of these selections are illustrated in Fig.~\ref{fig:qualitative_random}, and further examples are included in the supplementary material.

\begin{table}[t]
    \centering
    \caption{
        The \textbf{geometric error} of our method compared with RANSAC, ComplexGen, HPNet+Point2CAD and SEDNet+Point2CAD.
    }
    \begin{tabular}{l|r|r|r|r}
        \toprule
        \multirow{2}{*}{\textbf{Method}} & \multicolumn{3}{c|}{\textbf{CD ↓}} & \multirow{2}{*}{\textbf{LFD ↓}}                                    \\
                                         & \textbf{Vertex}                    & \textbf{Curve}                  & \textbf{Surface} &               \\\midrule
        \textbf{RANSAC Default}    & -                             & -                          & 0.0205           &  3123         \\
        \textbf{RANSAC Tuned}      & -                             & -                          & 0.0162           &  2502        \\
        \textbf{ComplexGen}              & 0.0901                             & 0.0601                          & 0.0402           & 4280          \\
        \textbf{HPNet+Point2CAD}         & 0.0782                             & 0.0222                          & 0.0157           & 2104          \\
        \textbf{SEDNet+Point2CAD}        & 0.0832                             & 0.0268                          & 0.0192           & 2262          \\
        \textbf{Ours}                & \textbf{0.0327}                   & \textbf{0.0144}                & \textbf{0.0093} & \textbf{908} \\
        \bottomrule
    \end{tabular}
    \label{table:geometric_error}
\end{table}

\begin{table}[t]
    \centering
    \caption{
        The \textbf{topological error} between the reconstructed and ground truth B-Rep models. 
        \textbf{FE} and \textbf{EV} denote the F1-Score of surface-curve connectivity and curve-vertex connectivity, respectively.
        Results from RANSAC are not available as they do not output topological structures.
    }
    \begin{tabular}{l|r|r}
        \toprule
        \textbf{Method}          & \textbf{FE ↑}    & \textbf{EV ↑}    \\\midrule
        \textbf{ComplexGen}      & 0.599          & 0.572          \\
        \textbf{HPNet+Point2CAD} & 0.739          & 0.674          \\
        \textbf{SEDNet+Point2CAD}   & 0.696          & 0.647          \\
        \textbf{Ours}            & \textbf{0.778} & \textbf{0.753} \\
        \bottomrule
    \end{tabular}
    \label{table:topological}
\end{table}

\begin{figure*}
    \centering
    \includegraphics[width=0.95\linewidth]{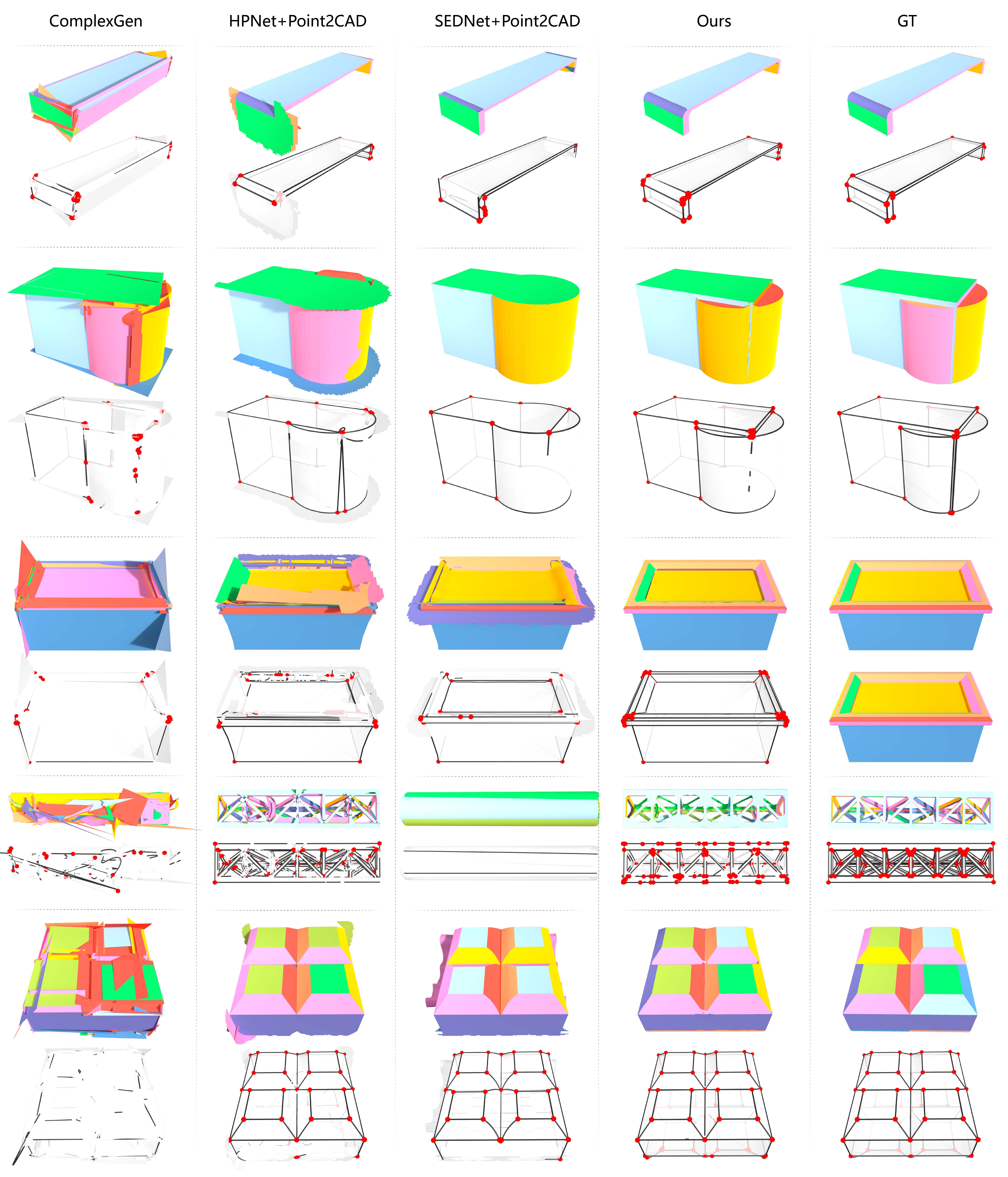}
    \caption{
        Qualitative comparisons of 5 \textbf{representative} shapes in ABC dataset.
    }
    \label{fig:qualitative_representative}
\end{figure*}

\begin{figure*}
    \centering
    \includegraphics[width=0.95\linewidth]{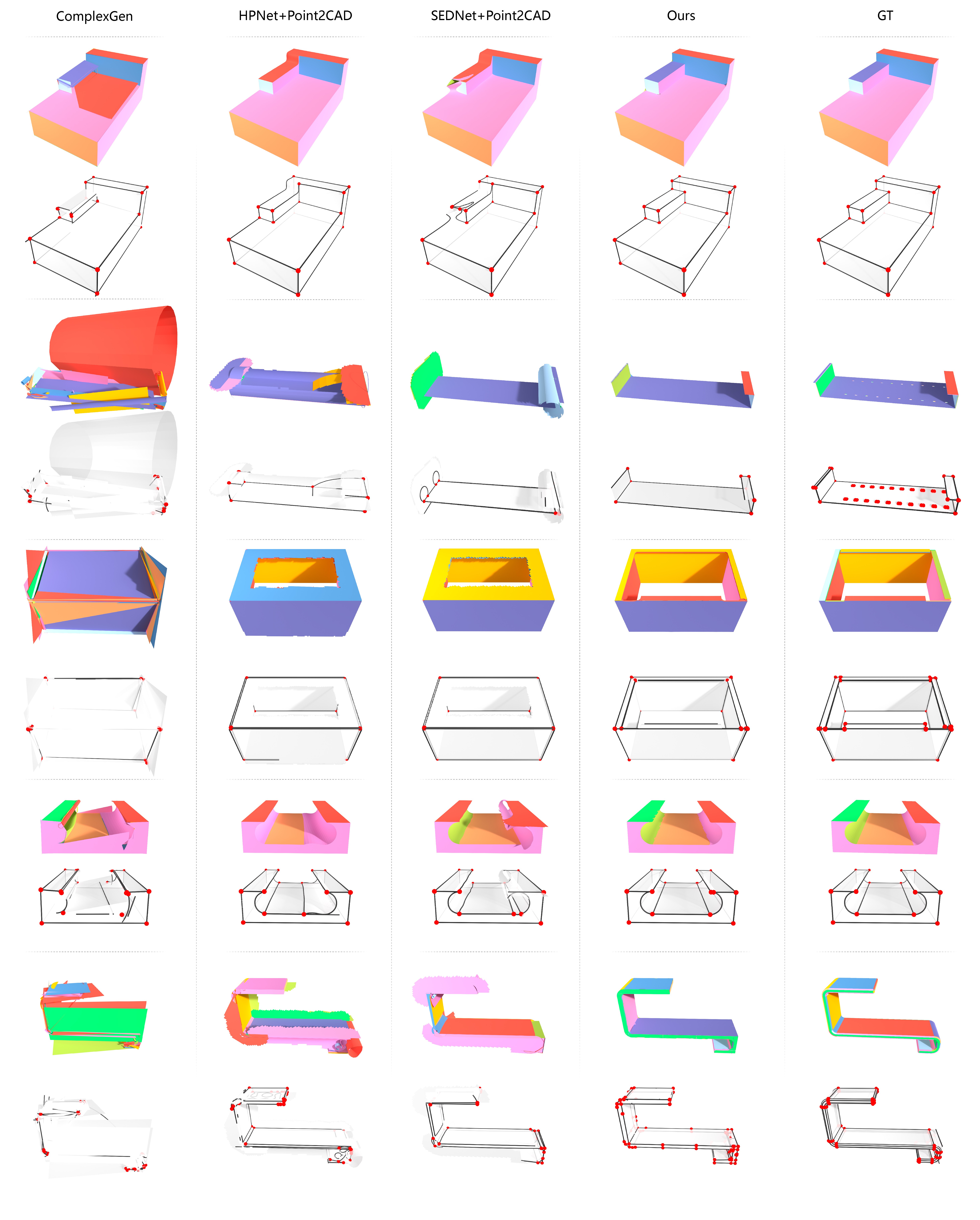}
    \caption{
        Qualitative comparisons of 5 \textbf{randomly sampled} shapes in ABC dataset.
    }
    \label{fig:qualitative_random}
\end{figure*}

\begin{table*}[t]
    \centering
    \caption{
        The \textbf{detection score} of vertices(V), curves(C) and surfaces(S) on the ABC dataset~\cite{ABC} using different methods.
        The evaluation process involved a matching process based on the Chamfer Distance, which was calculated between sample points on both the reconstructed and ground truth primitives.
        To ensure a thorough assessment, we employed a range of thresholds (0.1, 0.05, 0.02, 0.01, and 0.005).
        These multiple thresholds allowed for a more comprehensive evaluation in calculating Precision, Recall, and the F1-Score.
        With a chamfer distance threshold of 0.01, we additionally report the number of \textbf{good/total} primitives in the table.
    }
    \begin{tabular}{l|rrr|rrr|rrr|rrr}
        \toprule
        \multirow{2}{*}{\textbf{Method}} & \multicolumn{3}{c|}{\textbf{Number \#}} & \multicolumn{3}{c|}{\textbf{F1 Score ↑}} & \multicolumn{3}{c|}{\textbf{Precision ↑}} & \multicolumn{3}{c}{\textbf{Recall ↑}}                                                                                                                                         \\
                                         & \textbf{V}                              & \textbf{C}                               & \textbf{S}                                & \textbf{V}                            & \textbf{C}     & \textbf{S}     & \textbf{V}     & \textbf{C}     & \textbf{S}     & \textbf{V}     & \textbf{C}     & \textbf{S}     \\\midrule
        \textbf{RANSAC Default}    & -                                & -                                & 4.6/10                                 & -                                   & -          & 0.485          & -          & -          & 0.612          & -          & -          & 0.462          \\
        \textbf{RANSAC Tuned}      & -                                & -                                & 10.4/12.6                                 & -                                   & -          & 0.722          & -          & -          & 0.836          & -          & -          & 0.681          \\
        \textbf{ComplexGen}              & 7.9/22.8                                & 16.6/44.9                                & 10.5/25.9                                 & 0.5                                   & 0.513          & 0.502          & 0.521          & 0.510          & 0.479          & 0.505          & 0.538          & 0.552          \\
        \textbf{HPNet+Point2CAD}         & 17.7/26.3                               & 23.0/33.8                                & 8.9/12.8                                  & 0.671                                 & 0.696          & 0.697          & 0.749          & 0.755          & 0.776          & 0.649          & 0.668          & 0.661          \\
        \textbf{SEDNet+Point2CAD}        & 14.1/21.9                               & 17.1/26.6                                & 6.7/10.5                                  & 0.652                                 & 0.656          & 0.646          & 0.734          & 0.735          & 0.742          & 0.626          & 0.620          & 0.605          \\
        \textbf{Ours}                    & 29.6/38.5                               & 33.8/41.6                                & 14.2/15.8                                 & \textbf{0.785}                        & \textbf{0.790} & \textbf{0.821} & \textbf{0.810} & \textbf{0.850} & \textbf{0.902} & \textbf{0.802} & \textbf{0.766} & \textbf{0.781} \\
        \bottomrule
    \end{tabular}
    \label{table:area_detection}
\end{table*}

\begin{table}[t]
    \centering
    \caption{
        The \textbf{detection error} that directly computed on the input point cloud.
        Similar to Table~\ref{table:area_detection}, we compute the F1-Score, Precision and Recall using multiple thresholds (0.1, 0.05, 0.02, 0.01, 0.005).
    }
    \begin{tabular}{l|r|r|r}
        \toprule
        \textbf{Method}   & \textbf{F1 Score ↑} & \textbf{Precision ↑} & \textbf{Recall ↑} \\\midrule
        \textbf{HPNet}    & 0.614               & 0.670                & 0.589             \\
        \textbf{SEDNet}   & 0.588               & 0.674                & 0.554             \\
        \textbf{Ours w/o fitting} & \textbf{0.633}      & \textbf{0.711}       & \textbf{0.602}    \\
        \textbf{Ours w/ fitting} & \textbf{0.685}      & \textbf{0.757}       & \textbf{0.650}    \\
        \bottomrule
    \end{tabular}
    \label{table:point_detection}
\end{table}

\begin{figure}[t]
    \centering
    \includegraphics[width=0.99\linewidth]{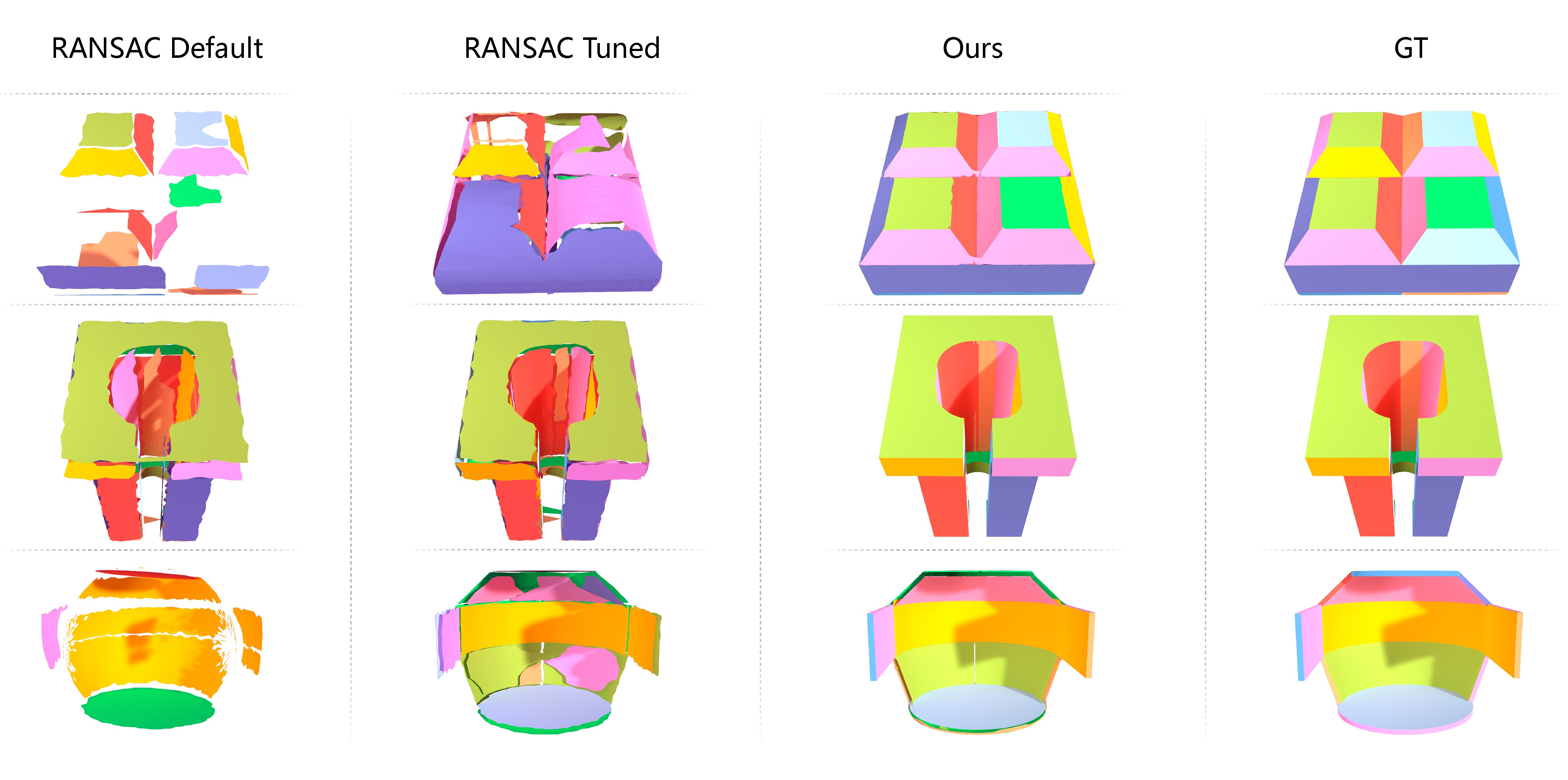}
    \caption{
        Qualitative comparisons with RANSAC~\cite{efficient_ransac}.
    }
    \label{fig:qua_ransac}
\end{figure}

As evidenced in Table~\ref{table:geometric_error}, our method surpasses all baselines in performance. 
The CAD models generated by our approach demonstrate superior accuracy and visually appeal. 
Additionally, Table~\ref{table:area_detection} indicates that our method outperforms others in detection metrics, excelling in both Precision and Recall.
Further, Table~\ref{table:topological} underscores our method's enhanced capability in capturing topological relations. 
These advantages are also visually apparent in Fig.~\ref{fig:qualitative_representative}, Fig.~\ref{fig:qualitative_random} and Fig.~\ref{fig:qua_ransac}.
We observe that detection-focused methods like ComplexGen struggle to accurately reconstruct CAD models, particularly for complex structures that diverge from their training datasets (details can be found in Sec.~\ref{sec:generalization}). 
Although HPNet and SEDNet demonstrate a higher degree of generalization, they are prone to inaccurately assigning points to primitives, often resulting in impractical primitive shapes. 
Furthermore, these methods often incorrectly predict the topological relationships between primitives. Such inaccuracies in primitive identification and topology typically lead to Point2CAD generating CAD models with fragmented and disconnected components.
Traditional methods such as RANSAC can accurately fit primitives but require finely tuned parameters. Moreover, they struggle to recover low-dimensional primitives like curves and vertices and fail to identify the topological relationships among the primitives.
In contrast, our method employs the Voronoi diagram as an intermediary representation.
The point assignment and the primitive fitting are separated into two individual processes, significantly enhancing both primitive reconstruction and topological accuracy. 

In our comprehensive analysis, we extend our evaluation to include a direct assessment of point clouds. 
This facilitates a more direct comparison of our method with two segmentation-based approaches, HPNet and SEDNet.
In Table~\ref{table:point_detection} we show two variants of our method, one with point labels directly segmented from our Voronoi Diagram (\textit{w/o fitting}) and the other with the point labels fitted to the extracted primitives (\textit{w/ fitting}).
Our method surpasses HPNet and SEDNet in key metrics such as F1-Score, Precision, and Recall. 
This further underscores the robustness and adaptability of our approach in diverse evaluation scenarios.

We also conducted a user study to assess the visual fidelity of CAD models reconstructed by different methods. For this study, we selected 20 shapes from our test set, which included 10 shapes from the representative shape set and another 10 from a set of randomly sampled shapes. 
Participants in the study were presented with the reconstructed CAD models from four different methods, along with their corresponding curves and vertices, and the ground truth for comparison.

\begin{figure}
    \centering
    \includegraphics[width=0.9\linewidth]{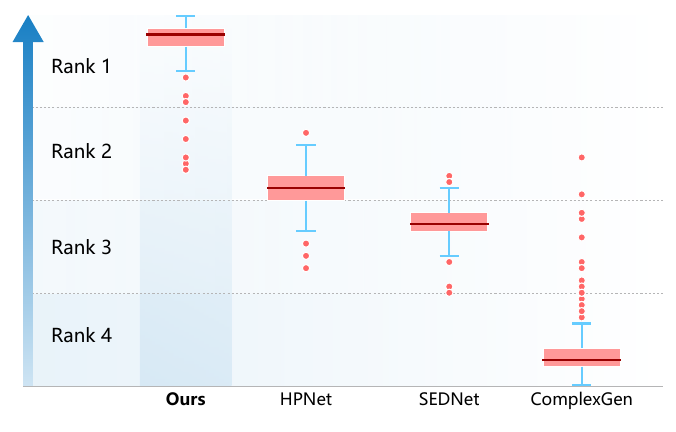}
    \caption{
        This is the box plot of our user study. On the Y-axis, we show the average ranking of each method.
    }
    \label{fig:user_study}
\end{figure}

The study involved 186 participants who were asked to evaluate and rank the models reconstructed by different methods based on their similarity to the ground truth. 
This approach allowed us to gauge user perceptions of the quality and accuracy of the reconstructed models. 
The results of this user study, which provide insights into the perceived quality of the models generated by each method, are depicted in Fig.~\ref{fig:user_study}.

The findings from the study clearly indicate that our method was ranked highest in terms of visual fidelity. 
This outcome underscores the effectiveness of our approach in producing CAD models that are not only accurate in terms of geometry and topology but also visually appealing to users.

\subsection{Generalization Ability}
\label{sec:generalization}
\begin{figure}[t]
    \centering
    \includegraphics[width=\linewidth]{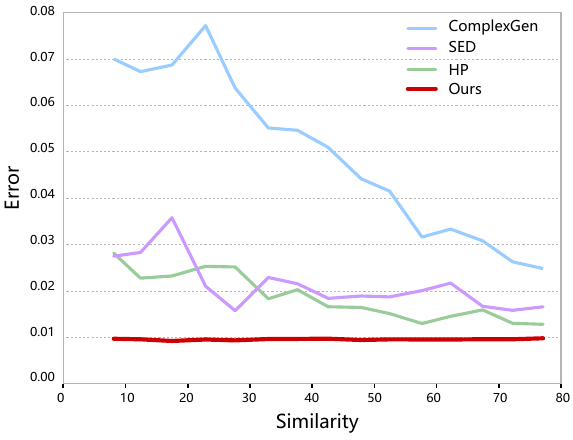}
    \caption{
        We plot the generalization ability of all methods on the test set.
        The similarity between the test model and the training set is sorted, and the corresponding reconstruction error is plotted.
        The similarity is reported in the range of 0\% to 100\%, where lower similarity indicates higher chamfer distance and less similarity to the training set.
        Our method has the best generalization ability of all the methods.
    }
    \label{fig:generalization}
\end{figure}

As mentioned above, ComplexGen~\cite{complexgen22} cannot faithfully reconstruct CAD models when the model is complex and less similar to the training set. 
We conducted a thorough analysis to investigate this issue further and explore the generalization capabilities of different methods.
We first identify the model in the training set that is most similar to each model in the test set. 
We utilized the Chamfer Distance between the corresponding models to quantify this similarity. 
A lower Chamfer Distance signifies a closer resemblance to the models in the training set, and a higher distance indicates lesser similarity.
We organized these similarity metrics in ascending order and plotted them alongside the corresponding geometric errors of each method. 
Fig.~\ref{fig:generalization} provides a clear illustration of how each method performs in terms of generalization across varying degrees of similarity to the training set.

\begin{table*}[t]
    \centering
    \caption{
        The performance of different methods under noisy input.
        We report the \textbf{geometric error} (CD), \textbf{detection score} (F1 score) and the \textbf{topological error} (F1 score) of different methods.
    }
    \begin{tabular}{l|rrr|rrr|rr}
        \toprule
        \multirow{2}{*}{\textbf{Method}} & \multicolumn{3}{c|}{\textbf{CD ↓}} & \multicolumn{3}{c|}{\textbf{Detection Score ↑}} & \multicolumn{2}{c}{\textbf{Topological Error ↑}}                                                                                         \\
                                         & \textbf{Vertex}                    & \textbf{Curve}                                  & \textbf{Surface}                                 & \textbf{Vertex} & \textbf{Curve} & \textbf{Surface} & \textbf{FE}    & \textbf{EV}    \\\midrule
        \textbf{ComplexGen}              & 0.1191                            & 0.0660                                         & 0.0442                                          & 0.460            & 0.478          & 0.462            & 0.546          & 0.520           \\
        \textbf{HPNet+Point2CAD}         & 0.1078                            & 0.0367                                         & 0.0235                                          & 0.445           & 0.492          & 0.542            & \textbf{0.722} & 0.657          \\
        \textbf{SEDNet+Point2CAD}        & 0.1044                            & 0.0352                                         & 0.0232                                          & 0.593           & 0.614          & 0.615            & 0.685          & 0.629          \\
        \textbf{Ours}                    & \textbf{0.0539}                   & \textbf{0.0294}                                & \textbf{0.0121}                                 & \textbf{0.683}  & \textbf{0.683} & \textbf{0.742}   & 0.707          & \textbf{0.673} \\
        \bottomrule
    \end{tabular}
    \label{table:noise}
\end{table*}

As demonstrated in Fig.~\ref{fig:generalization}, ComplexGen's performance degrades notably with diminishing shape similarity to the training set, evidenced by an increased reconstruction error. 
The full information change between the voxel and primitive features in ComplexGen makes the learning process ambiguous, impairing its generalization capacity.
In stark contrast, our method maintains consistent performance, irrespective of the model's similarity to the training dataset. This stability is due, in large part, to our method's approach to the Voronoi diagram prediction, which simplifies the task to a binary classification problem. 
This simplification renders our method significantly more robust compared to ComplexGen's mixed combinatorial and continuous learning process.
Furthermore, our approach to predicting the Voronoi diagram relies primarily on local geometric cues, as detailed in Sec.~\ref{sec:method_voronoi} 
This focus on local geometry fosters a more generalizable learning process, enabling our method to maintain high accuracy and reliability across a wider range of shapes and complexities. 
This attribute starkly distinguishes our method from others, particularly in scenarios involving diverse and complex CAD models.

Interestingly, we even have a slightly higher error on the most similar shapes (70\%-80\%).
We attribute this to the metric we used to measure the similarity.
Chamfer distance from the test shape to training items is not the most ideal choice. 
For instance, some shapes might have a small Chamfer distance not because they are similar but rather because they share a similar size (e.g., two cubes of differing sizes might have a large chamfer distance even though they have similar appearance).

\begin{figure*}
    \centering
    \includegraphics[width=0.7\linewidth]{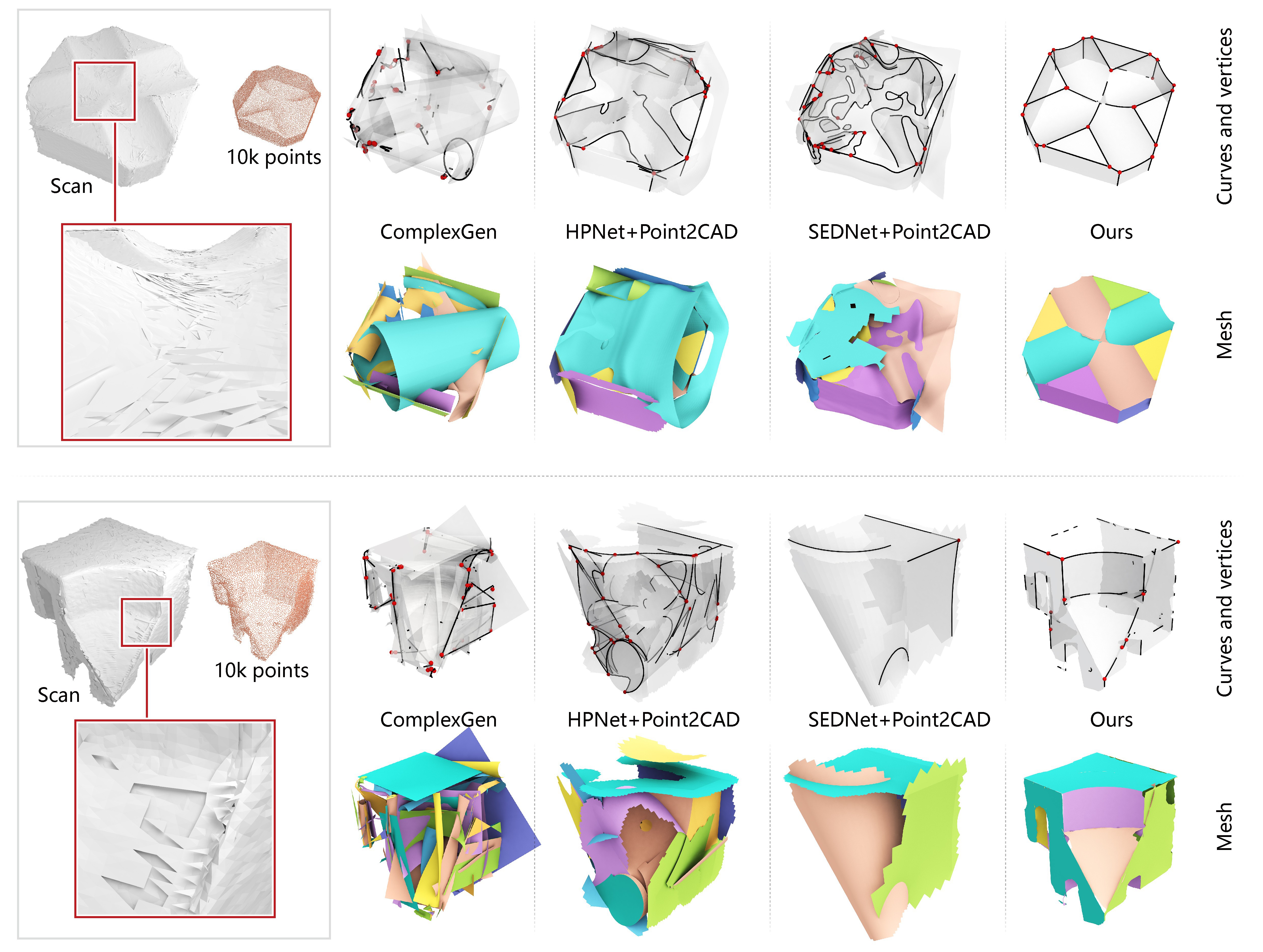}
    \caption{
        Qualitative comparison on two real scans. 
        The input models are derived from structured light scanners~\cite{DEF22}. 
        These scans are characterized by significant noise and data omissions, including issues like noise accumulation during scanning, self-intersecting triangles, and errors in alignment leading to duplication and overlapping. 
        We standardized the input to 10,000 points for all methods to ensure a fair comparison. 
        The pronounced noise particularly challenges previous methods' mixed combinatorial and continuous learning processes, resulting in suboptimal segmentation and inaccurate geometric parameters. 
        Despite being trained predominantly on synthetic data, our method demonstrates a superior capacity to manage and interpret noisy data.
    }
    \label{fig:qualitative_real}
\end{figure*}

\begin{figure}
    \centering
    \includegraphics[width=0.9\linewidth]{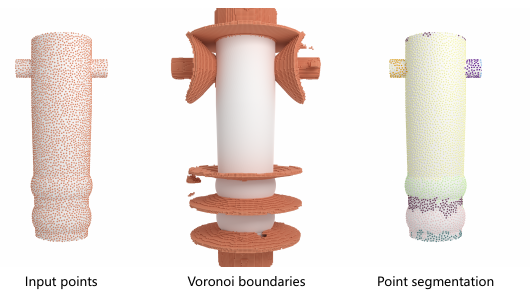}
    \caption{
        Although we omitted the BSpline primitive in our mesh extraction process, 
        NVD can still faithfully predict the Voronoi boundaries for \textbf{BSpline primitives} and subsequently assign the instance label to each point.
        We show the input points (\textit{left}), the predicted Voronoi boundaries (\textit{middle}) and the corresponding point segmentation (\textit{right}).
    }
    \label{fig:bspline}
\end{figure}

\subsection{Stress Test}
We extended our evaluation to more challenging scenarios, including noisy point clouds and real-world scans.
For the noisy point cloud tests, we introduced random noise equivalent to 1\% of each shape's diagonal length~\cite{sed23}. 
We directly use the pre-trained models of all methods to reconstruct the noisy shapes.
The performance of each method in handling these noisy inputs is presented in Table~\ref{table:noise}. 
Our findings indicate that our method still leads in terms of geometric error and detection score, while maintaining comparable topological performance to other methods. 
This highlights our method's resilience and accuracy even in the presence of data imperfections.

Regarding real scans, we employed meshes reconstructed from structured light scanners as described in DEF~\cite{DEF22}. 
This allowed us to test our method's effectiveness on data derived from real-world objects, further extending its applicability. 
The outcomes of these tests, which showcase the capability of our method to handle real scan data, are visually depicted in Fig.~\ref{fig:qualitative_real}. 
This visual representation underscores our method's practical utility and robustness in diverse and challenging environments, beyond the controlled conditions of synthetic datasets.

\begin{figure*}[t!]
    \centering
    \includegraphics[width=\linewidth]{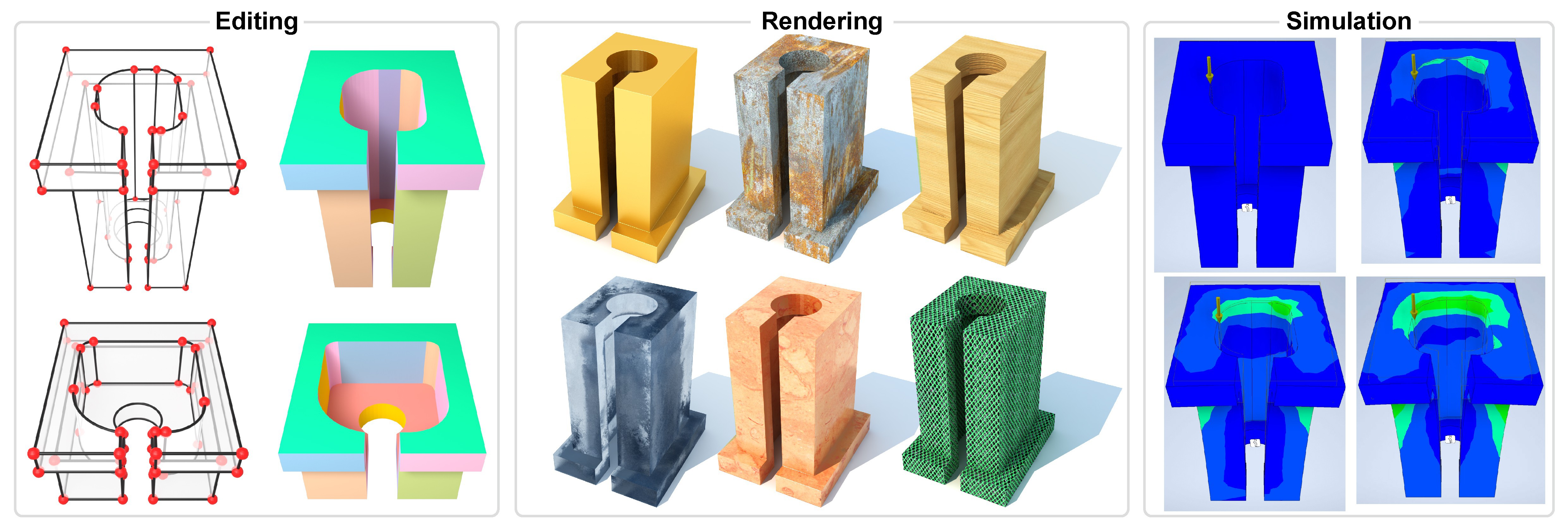}
    \caption{
        Three distinct applications utilizing the reconstructed B-Rep model.
        The reconstructed B-Rep model's compactness, cleanliness, and preservation of sharp features make them exceptionally well-suited for various subsequent uses, including editing, rendering, physical simulation, etc.  
    }
    \label{fig:app}
\end{figure*}

\subsection{Limitation}
\label{sec:limitation}
Although our method achieves state-of-the-art performance in CAD model reconstruction, some limitations remain.

\paragraph{Voxelized Voronoi Diagram:}
The voxelized representation of the Voronoi diagram introduces inherent limitations, such as the resolution constraints of the voxel grid. 
This can particularly impact thin primitives: in cases where Voronoi boundaries entirely occupy a primitive's Voronoi cell, both the cell and the corresponding primitive are omitted in subsequent processes. 
This issue is evident in the first shape of Fig.~\ref{fig:qualitative_random}, where a thin plane fails to be recovered, resulting in a zigzag boundary of the yellow nearby plane. 
Additionally, small holes in the predicted Voronoi boundaries may inadvertently connect Voronoi cells, leading to suboptimal primitive fitting (as shown in Fig.~\ref{fig:bspline}).

\paragraph{Noise data:}
Fig.~\ref{fig:qualitative_real} illustrates that while our method can reconstruct a reasonably accurate B-Rep model to a certain extent, the process of CAD model reconstruction in the presence of noise remains challenging. The complexity stems from the non-uniform nature of noise distribution, which is influenced by a variety of factors including the characteristics of the acquisition device, environmental conditions, and the specifics of the reconstruction algorithm. This diversity in noise sources and patterns makes it challenging to model the noise distribution, thereby complicating the training of a universally robust model capable of handling all types of noise.

\paragraph{B-Spline Fitting:}
While advancements have been made in fitting B-Spline primitives from unstructured point clouds~\cite{bspline12,parsenet20,point2cad}, the process remains unstable and occasionally yields implausible results. 
The intersection of B-Spline primitives adds further complexity to this process. 
Consequently, our experiments primarily utilized elementary primitives. 
However, B-Spline primitives were still employed in training the network for Voronoi boundary prediction, given that Voronoi diagram construction is independent of primitive type. 
We also visualize our segmentation results for B-Spline primitives in Fig.~\ref{fig:bspline}.

\paragraph{Meshing:}
Despite the ability of current methods, including ComplexGen~\cite{complexgen22}, HPNet~\cite{hpnet21}, SEDNet~\cite{sed23}, and ours, to accurately determine primitives and their topologies, the meshing process remains erratic. 
A key challenge is that the primitives are topologically but not geometrically interconnected, affecting loop finding and trimming processes. 
ComplexGen attempts to address this through an optimization framework that adjusts geometric and topological relations simultaneously, but it still struggles with the instability of meshing. 
Developing a stable meshing process remains an unresolved issue in this field.

\paragraph{Shapes violate piece-wise G2 continuity:}
Our method utilizes training data generated from CAD models that adhere to piece-wise G2 continuity, with discontinuities occurring only at the junctions between primitives. 
When encountering input shapes that deviate from this continuity, such as partial inputs or surfaces with G2 continuous junctions, our method might struggle to predict Voronoi boundaries accurately. 
Nevertheless, it can effectively reconstruct the Voronoi boundaries and primitives in the intact regions of partial inputs, thanks to local feature learning (as shown in Fig.~\ref{fig:qualitative_real}). 
In cases of surfaces with G2 continuous transitions, the reconstruction leans towards semantic analysis over geometric detail, necessitating a broader understanding of the shape's features. 
To improve performance on such inputs, retraining NVD-Net with a dataset inclusive of shapes with G2 continuous transitions could be beneficial, as these also possess well-defined Voronoi Diagrams.

\section{Conclusion and Future Work}
\label{sec:conclusion}

In this work, we introduce a novel pipeline for reconstructing Boundary Representation (B-Rep) CAD models, capable of processing diverse inputs such as point clouds, distance fields, and meshes. 
The enabler of our approach is the prediction of the Voronoi diagram representing the underlying shape. 
It serves as a foundation for extracting the geometric primitives and their topological relationships, which are then integrated to reconstruct the CAD models in B-Rep. 
The Voronoi diagram, being both \textit{unique} and \textit{fixed}, significantly reduces ambiguity in the reconstruction process. 
Its inherent structure conveniently encapsulates boundaries and connectivity of primitives, thereby simplifying the reconstruction workflow. 
Through extensive validation, our method has demonstrated superior performance over existing techniques in geometric accuracy, detection precision, topological consistency, and generalizability.
The reconstructed B-Rep models are compatible with existing CAD software, enabling seamless integration into the design process; see Fig.~\ref{fig:app}.

The main limitation of our method is the accuracy of the Voronoi diagram. 
The voxelization process inherent to our method introduces quantization errors that adversely impact the Voronoi diagram's precision. 
This effect is particularly pronounced in the case of tiny structures, which may be lost during voxelization and subsequently affect the extraction of primitives. 
Future research will investigate alternative representations of the Voronoi diagram, such as implicit functions, to circumvent the quantization errors associated with voxelization.

\section*{Acknowledgments}
We thank all the anonymous reviewers for their insightful comments. Thanks also go to Haoxiang Guo, Dani Lischinski, and Ningna Wang for helpful discussions. This work was supported in part by NSFC (62161146005, U21B2023, U2001206), Guangdong Basic and Applied Basic Research Foundation (2023B1515120026), DEGP Innovation Team (2022KCXTD025), Shenzhen Science and Technology Program (KQTD20210811090044003, RCJC20200714114435012, JCYJ20210324120213036), NSERC (611370), ISF (3441/21), SFU Graduate Dean's Entrance Scholarship, Guangdong Laboratory of Artificial Intelligence and Digital Economy (SZ), and Scientific Development Funds from Shenzhen University.

\bibliographystyle{ACM-Reference-Format}
\bibliography{NeuralVoronoi}
 
\clearpage
\appendix
\title{Split-and-Fit: Learning B-Reps via Structure-Aware Voronoi Partitioning (Supplementary Materials)}
\maketitlesup
This supplementary material presents the training details of NVD-Net and a pseudocode describing the subsequent shape extraction algorithm.
We also include a qualitative comparison of four methods on 30 representative shapes and 30 randomly sample shapes in the test set (Sec.~\ref{sec:additional_results}).
  
\section{Primitive Extraction}

Our primitive extraction method begins by segmenting the input shape into Voronoi cells, aiming to simplify the complex task of primitive fitting by localizing the process within each cell, see Sec.~\ref{sec:hole}. This segmentation facilitates a more manageable and efficient primitive fitting process as each cell typically contains fewer primitives than the entire shape, significantly improving performance over traditional global fitting methods like RANSAC~\cite{ransac,efficient_ransac}. 
Similar to SEDNet~\cite{sed23} and Point2CAD~\cite{point2cad}, we additionally recover curves and vertices through primitive intersections. We show the pseudocode of our primitive extraction in Algorithm~\ref{alg:primitive}. The process culminates in the generation of a comprehensive B-Rep model, detailed through vertices (V), curves (E), and surfaces (F), bound together by their geometric and topological relationships, see Sec.~\ref{sec:fitting}.

\begin{table}[t]
\caption{Summary of important notations.}
\label{tab:notions}
\centering
\begin{tabularx}{\columnwidth}{@{}rX@{}} 
\toprule
    Notation & Description \\
    \midrule
    $\mathbb{B}$ & Boolean domain $\{0,1\}$ \\
    $N_{v/e/f}$ & element numbers for vertices/curves/surfaces \\
    $M = (V, E, F, \partial, \mathcal{P})$ & a B-Rep chain complex with vertices $V$, curves $E$, surfaces $F$, topological embedding $\partial=\{FF,FE,EE,EV,FV\}$, and geometric embedding $\mathcal{P}$ \\
    $FF \in \mathbb{B}^{N_f \times N_f}$ & the adjacency of surfaces and surfaces \\
    $FE \in \mathbb{B}^{N_f \times N_e}$ & the adjacency of surfaces and curves \\
    $EE \in \mathbb{B}^{N_e \times N_e}$ & the adjacency of curves and curves \\
    $EV \in \mathbb{B}^{N_e \times N_v}$ & the adjacency of curves and vertices \\
    $FV \in \mathbb{B}^{N_f \times N_v}$ & the adjacency of surfaces and vertices \\
    $P_i$ & the internal points in each Voronoi cell \\
    $\epsilon_1$ & 0.001, threshold of Fitting\_Primitives \\
    $\epsilon_2$ & 0.02, threshold of Build\_Surfaces\_Adjacency \\
    $\epsilon_3$ & 0.05, threshold of Build\_Curves\_Adjacency \\
\bottomrule
\end{tabularx}
\end{table}

\begin{algorithm}[t]
    \SetCommentSty{commfont}
    \SetKwInOut{Input}{input}\SetKwInOut{Output}{output}
    
    \Input{Voronoi cells $N_v$}
    \Output{B-Rep model $M(V,E,F,\partial,\mathcal{P})$} \BlankLine
    {$V,E,F \leftarrow \emptyset$}\;
    {SType $\leftarrow$ \{plane, sphere, cylinder, cone, torus\}}\;
    {CType $\leftarrow$ \{line, circle, ellipse\}}\;

    $F,E \leftarrow$ Fitting\_Primitives($N_v$)\;
    $\textbf{FF} \leftarrow$ Build\_Surfaces\_Adjacency($N_v,F$)\;
    $E, \textbf{FE} \leftarrow$ Curve\_Extraction($F,E,\textbf{FF}$)\;
    $\textbf{EE} \leftarrow$ Build\_Curves\_Adjacency($F,E,\textbf{FE}$)\;
    $V, \textbf{EV}, \textbf{FV} \leftarrow$ Vertex\_Extraction($E,\textbf{EE}$)\;
            
    \caption{Overview of our primitive extraction process}
    \label{alg:primitive}
\end{algorithm}

\begin{algorithm}[t]
    \SetCommentSty{commfont}
    \SetKwInOut{Input}{input}\SetKwInOut{Output}{output}
    \Input{Voronoi cells $N_v$}
    \Output{Surfaces $F$ and Curves $E$} 
    \BlankLine
    {$E,F \leftarrow \emptyset$}\;
    
    Error $\epsilon \leftarrow \inf$\;
    \For{each $N_i$ in $N_v$ }{
        Surface $f$, $\epsilon \leftarrow$ Plane\_Fitting($N_i$)\;
        \If{$\epsilon < \epsilon_1$}
        {
            \tcp{find the best curve in CType using points in $N_i$}
            Curve $e$, $\epsilon \leftarrow$ Curve\_Fitting($N_i$)\;
            \If{$\epsilon < \epsilon_1$}
            {
                $E \leftarrow E \cup e$\;
                continue\;
            }
            \Else{
                $F \leftarrow F \cup f$\;
            }
        }
        \Else{
            \tcp{find the best surface in SType using points in $N_i$}
            Surface $f$, $\epsilon \leftarrow$ Surface\_Fitting($N_i$)\;
            \If{$\epsilon < \epsilon_1$}
            {
                $F \leftarrow F \cup f$\;
            }
            \Else
            {
                \tcp{Use RANSAC to fit multiple primitives}
                Surface set $F_s \leftarrow$ RANSAC($N_v$)\;
                $F \leftarrow F \cup F_s$\;
            } 
        }
    }
    \caption{Fitting\_Primitives()}
    \label{alg:fitting}
\end{algorithm}

\begin{algorithm}[t]
    \SetCommentSty{commfont}
    \SetKwInOut{Input}{input}\SetKwInOut{Output}{output}
    
    \Input{Voronoi cells $N_v$ and Surfaces $F$}
    \Output{Surfaces adjacency $\textbf{FF}$} 
    $\textbf{FF}[N_f][N_f] \gets 0$\;
    \For{each $N_i$ in $N_v$}{
        \For{each $N_j$ in adjacent cells}{
            \If{$is\_SType(N_i)$ and $is\_SType(N_j)$}
            {
                $\epsilon \leftarrow Distance(P_i,P_j)$\;
                \If{$\epsilon < \epsilon_2$}
                {
                    $\textbf{FF}[i][j] \leftarrow 1$\;
                }
            }
        }
    }
    \caption{Build\_Surfaces\_Adjacency()}
    \label{alg:ff-adj}
\end{algorithm}

\begin{algorithm}[t]
    \SetCommentSty{commfont}
    \SetKwInOut{Input}{input}\SetKwInOut{Output}{output}
    
    \Input{Surfaces $F$, Curves $E$ and Surfaces adjacency $\textbf{FF}$}
    \Output{Curves $E$ and Surface-Curve adjacency $\textbf{FE}$} 

    \For{each $f_i$ in $F$}{
        \For{each $f_j$ in adjacent surfaces}{
            Common points $P \leftarrow P_i \cap P_j$\;
            Compute intersection curves $E_s$\;
            \tcp{find the curve with minimal distance to $P$}
            Curve $e \leftarrow Min\_Distance(E_s,P)$\;
            $E \leftarrow E \cup e$\;
            update Surface-Curve adjacency $\textbf{FE}$\;
        }
    }
    {remove duplication in $E$ and update $\textbf{FE}$}\;
    \caption{Curve\_Extraction()}
    \label{alg:ff-inter}
\end{algorithm}

\begin{algorithm}[t]
    \SetCommentSty{commfont}
    \SetKwInOut{Input}{input}\SetKwInOut{Output}{output}
    
    \Input{Surfaces $F$, Curves $E$ and Surfaces adjacency $\textbf{FE}$}
    \Output{Curves adjacency $\textbf{EE}$} 
    $\textbf{EE}[N_e][N_e] \gets 0$\;
    \For{each $e_i$ in $E$}{
        \For{each $e_j$ in $E$}{
            \For{each $f_k$ in $F$}{
                \If{$\textbf{FE}[k][i] = 1$ and $\textbf{FE}[k][j] = 1$}
                {
                    $\epsilon \leftarrow Distance(P_i,P_j)$\;
                    \If{$\epsilon < \epsilon_3$}
                    {
                        $\textbf{EE}[i][j] \leftarrow 1$\;
                    }
                }
            }
        }
    }
    \caption{Build\_Curves\_Adjacency()}
    \label{alg:cc-adj}
\end{algorithm}

\begin{algorithm}[t]
    \SetCommentSty{commfont}
    \SetKwInOut{Input}{input}\SetKwInOut{Output}{output}
    
    \Input{Curves $E$ and Curves adjacency $\textbf{EE}$}
    \Output{Vertices $V$, Curve-Vertex adjacency $\textbf{EV}$ and Surface-Vertex adjacency $\textbf{FV}$} 

    \For{each $e_i$ in $E$}{
        \For{each $e_j$ in adjacent curves}{
            Compute intersection vertices $V_s$\;
            \tcp{find the vertex with minimal distance to $e_i$ and $e_j$}
            Vertex $v \leftarrow Min\_Distance(V_s, e_i, e_j)$\;
            $V \leftarrow V \cup v$\;
            update Curve-Vertex adjacency $\textbf{EV}$\;
            update Surface-Vertex adjacency $\textbf{FV}$\;
        }
    }
    {remove duplication in $V$ and update $\textbf{EV,FV}$}\;
    \caption{Vertex\_Extraction()}
    \label{alg:cc-inter}
\end{algorithm}

\subsection{Hole Filling and Cell Construction}
\label{sec:hole}
We identify Voronoi cells from the predicted Voronoi boundaries using region growing. 
Due to holes commonly appearing in these boundaries, we perform hole-filling to refine them before reconstructing the cells. 
This involves traversing each voxel in the space along the gradient direction to gather a set of voxels and their classification results. 
If half or more of these are part of the Voronoi boundary, we also classify the rest as boundary cells. 
With these refined boundaries, we apply standard region growing to group the voxels into distinct Voronoi cells.

\subsection{Primitive Fitting}
\label{sec:fitting}
We provide an overview of our primitive fitting process in Algorithm~\ref{alg:primitive} and the chosen parameters in Table.~\ref{tab:notions}.
We first define two distinct sets of geometric primitives. 
For curves, our consideration extends to lines, circles, and ellipses, categorized under \textit{CType}; 
for surfaces, we encompass planes, spheres, cylinders, cones, and tori, collectively referred to as \textit{SType}.
We use least-square fitting to obtain all these primitives \footnote{\url{https://www.geometrictools.com/Samples/Mathematics.html}}.
Details of the primitive fitting process are described as follows (Algorithm~\ref{alg:fitting}):
\begin{itemize}
    \item \textbf{Sequential Fitting}: We perform a sequential fitting process for each Voronoi cell. Initially, we attempt to fit the simplest primitive, plane, and evaluate the fitting error using a predefined threshold $\epsilon_1$. 
    Since a cell of a curve may also be recognized as a plane, we proceed to curve fitting within the cell if the plane is successfully fitted, trying each curve type in \textit{CType} and selecting the one with the minimal fitting error. 
    If no satisfactory curve fitting is achieved (minimal fitting error > $\epsilon_1$), we explore surface fitting by iterating through \textit{SType}. 
    A similar error-based selection criterion is applied to determine the best-fitting surface.
    \item \textbf{Fallback Mechanism}: For cells where a single primitive fitting is unfeasible, we employ a traditional RANSAC algorithm to detect combinations of primitives, ensuring robustness in handling complex geometries.
\end{itemize}

Once we have done the primitive fitting, subsequent steps delve into defining topological relationships and extracting finer geometric details.  
Details of each step are described as follows:
\begin{itemize}
    \item \textbf{Surfaces Adjacency (Algorithm ~\ref{alg:ff-adj})}: We consider surfaces to be adjacent if they belong to neighbouring Voronoi cells and the distance between their points is less than a predefined threshold $\epsilon_2$. 
    This adjacency relationship is stored in the surfaces adjacency matrix (\textbf{FF}). This dual condition ensures a robust method of determining adjacency, where spatial proximity and Voronoi adjacency work in tandem. 
    \item \textbf{Curve Extraction (Algorithm ~\ref{alg:ff-inter})}: The intersection of surfaces is computed to refine the curve elements (E)\footnote{\url{https://dev.opencascade.org/doc/overview/html/occt_user_guides__modeling_algos.html}}. Since two surfaces may intersect along multiple curves, we identify the correct boundary curve by finding common points shared by the surfaces. This process ensures that the extracted curves precisely represent the shape's edges in the B-Rep model.
    \item \textbf{Curves Adjacency (Algorithm ~\ref{alg:cc-adj})}: We define curves as adjacent if they are connected to the same surface and the distance between their internal points is less than a predefined threshold $\epsilon_3$. This adjacency relationship is stored in the curves adjacency matrix (\textbf{EE}). 
    \item \textbf{Vertex Extraction (Algorithm ~\ref{alg:cc-inter})}: Similar to the curve extraction, we extract vertices (V) through the intersection of curves, completing the boundary representation (B-Rep) model construction.
\end{itemize}

\section{Network and Training Details}
\paragraph{Network: }
The backbone of the approach is a simple UNet architecture, which comprises four down-sampling and four up-sampling layers. 
The architecture is detailed as follows:
\begin{equation}
    \begin{split}
        & Conv(1, 4, 16)-BatchNorm-ReLU-\\
        & DownBlock(16,32)-DownBlock(32,64)- \\
        & DownBlock(64,128)-DownBlock(128,256)- \\
        & UpBlock(256,128)-Block(256,128)- \\
        & UpBlock(128,64)-Block(128,64)- \\
        & UpBlock(64,32)-Block(64,32)- \\
        & UpBlock(32,16)-Block(32,16)-Block(16,1)-Sigmoid, \nonumber
    \end{split}
\end{equation}
where $Conv(kernel size, in, out)$ denotes a 3D convolutional layer.
$Block$ consists of a $Conv(3, in, out)$, a $BatchNorm$, a $ReLU$ layer, a $Conv(3, out, out)$, a $BatchNorm$ and a $ReLU$ layer.
$DownBlock$ consists of a $Block(in, out)$ and a $MaxPool$ layer.
$UpBlock$ consists of a $Upsample$, a $Conv(3, in, out)$ and a $ReLU$ layer.
All the $UpBlock$ has a skip concatenation from the corresponding $DownBlock$ layer.

\paragraph{Data: }
For training data, we use the same split as the previous
work~\cite{hpnet21,sed23,complexgen22}. We use the shape number from 0-800000
in ABC dataset for training, 800000-900000 for validation and 900000-1000000
for testing. For each shape, we use Poisson disk sampling to sample 10000
points and feed it into NDC~\cite{ndc22} to generate the UDF field and
corresponding gradient direction as the network input. As for the ground
truth label of the Voronoi diagram, we first voxelize the unit square bounding box
of the shape into a $256^3$ grid. Based on the annotation provided by the ABC
dataset, we compute and compare the nearest primitive for each voxel. By
definition, if two adjacent voxels have different nearest primitives, there is
a Voronoi boundary between them. We label both voxels as the voxelized Voronoi
boundaries and train the network to predict this label. To learn local
features, we split the voxel grid into $32^3$ local patches and train the
network to predict the Voronoi boundaries for each local patch. We also filter
out patches far from the actual shape (i.e., the distance to the
shape is larger than 0.3).

\paragraph{Training: }
We use Adam optimizer with a learning rate 0.0001 and a batch size 16.
The weight of focal loss is set to 0.75 since most of the voxels have
negative labels. We also augment the data by randomly flipping the input UDF field
and the corresponding ground truth label.

\section{Additional Results}
\label{sec:additional_results}
\begin{figure*}
    \centering
    \includegraphics[width=0.95\linewidth]{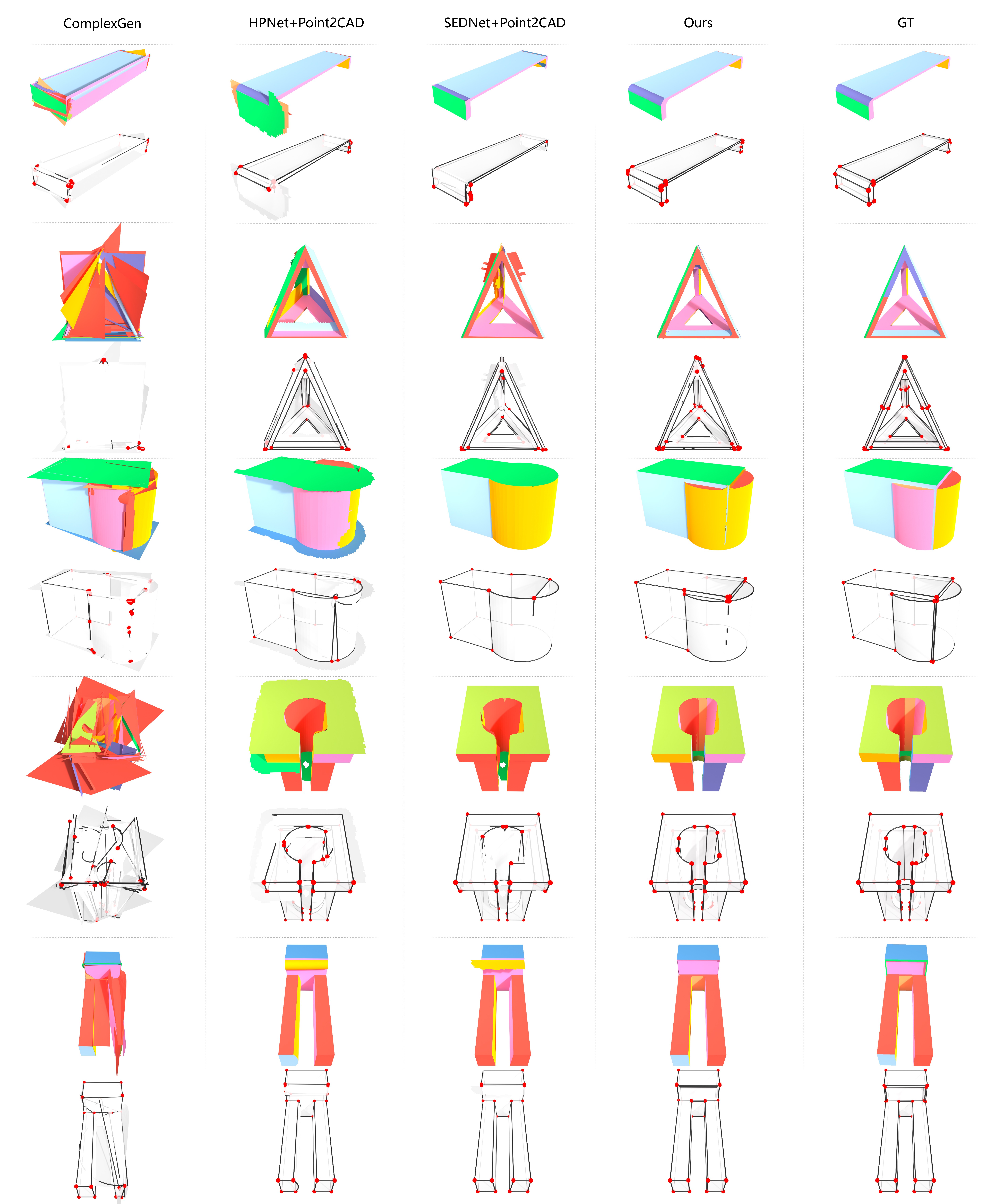}
    \caption{
        Qualitative comparisons of \textbf{representative} cases (1/6).
    }
    \label{fig:vis31_1}
\end{figure*}

\begin{figure*}
    \centering
    \includegraphics[width=0.95\linewidth]{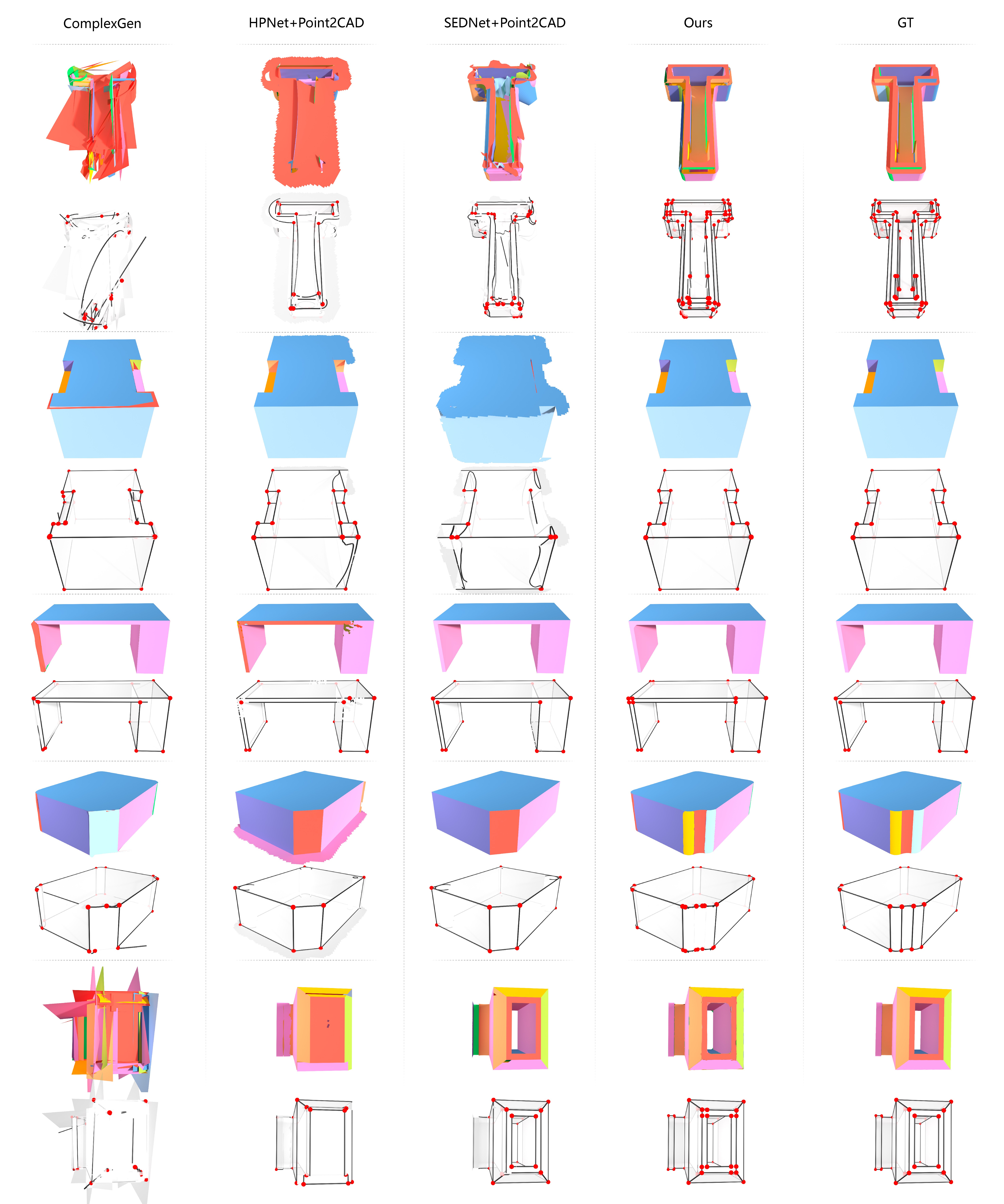}
    \caption{
        Qualitative comparisons of \textbf{representative} cases (2/6).
    }
    \label{fig:vis31_2}
\end{figure*}

\begin{figure*}
    \centering
    \includegraphics[width=0.95\linewidth]{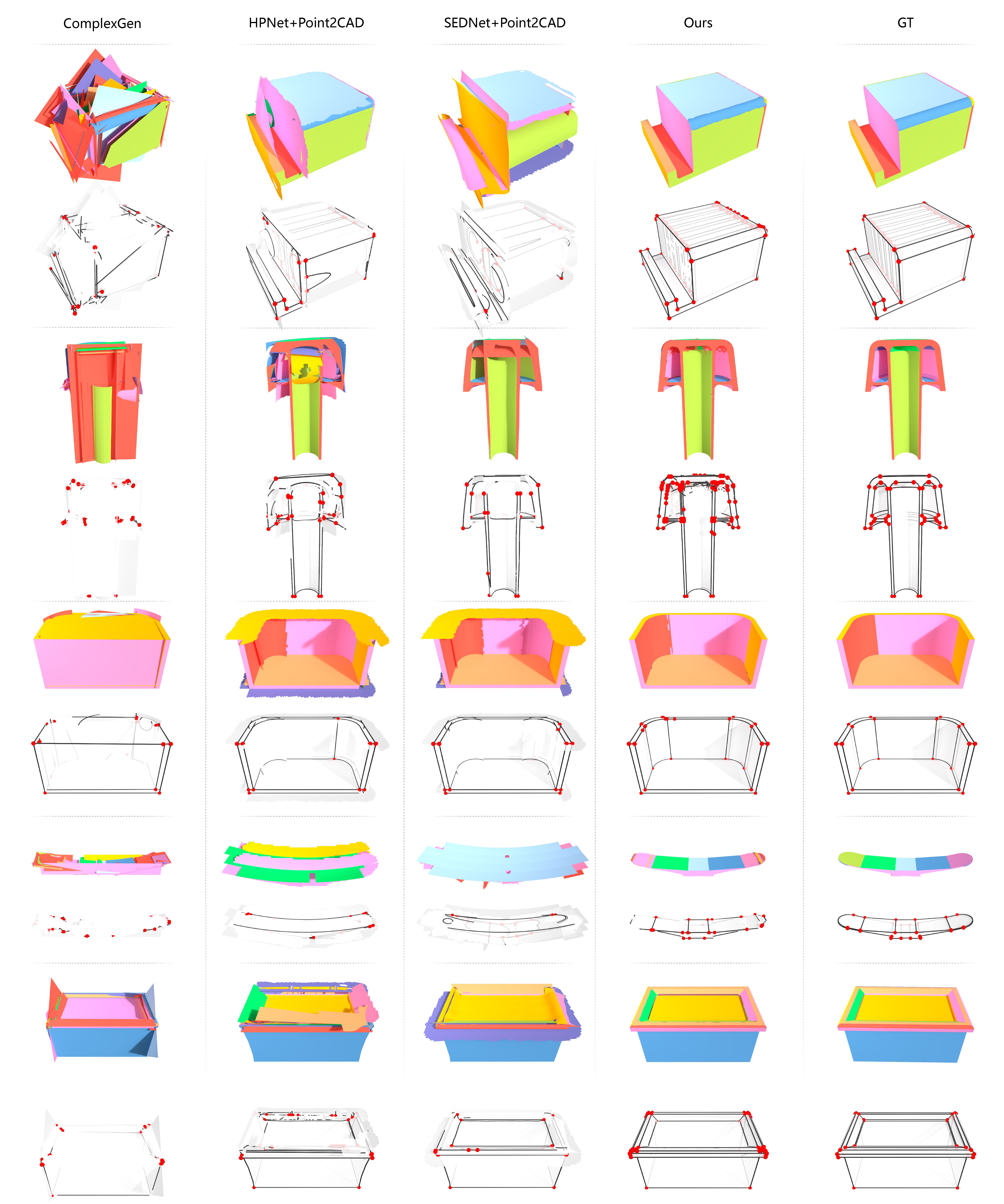}
    \caption{
        Qualitative comparisons of \textbf{representative} cases (3/6).
    }
    \label{fig:vis31_3}
\end{figure*}

\begin{figure*}
    \centering
    \includegraphics[width=0.95\linewidth]{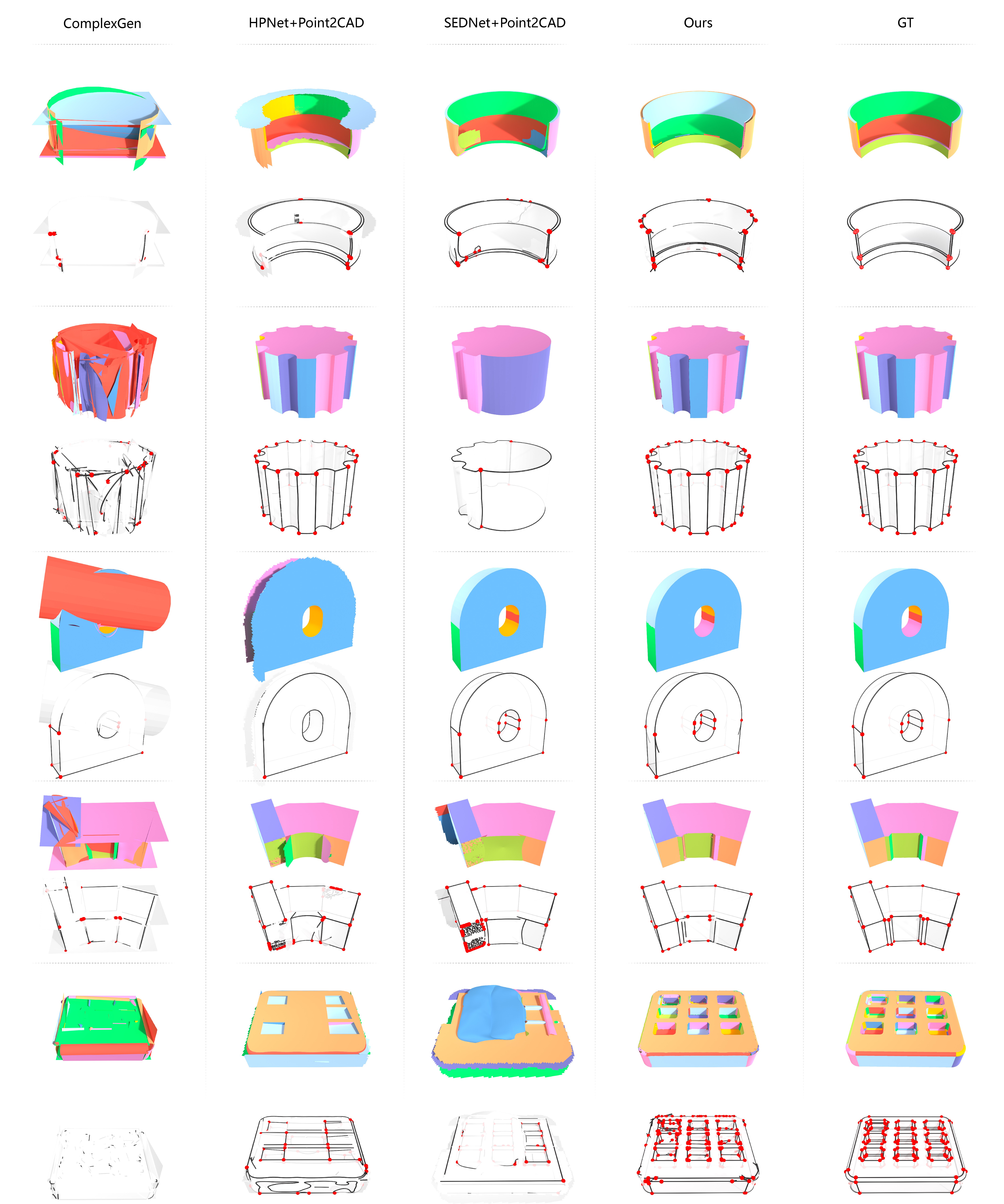}
    \caption{
        Qualitative comparisons of \textbf{representative} cases (4/6).
    }
    \label{fig:vis31_4}
\end{figure*}

\begin{figure*}
    \centering
    \includegraphics[width=0.95\linewidth]{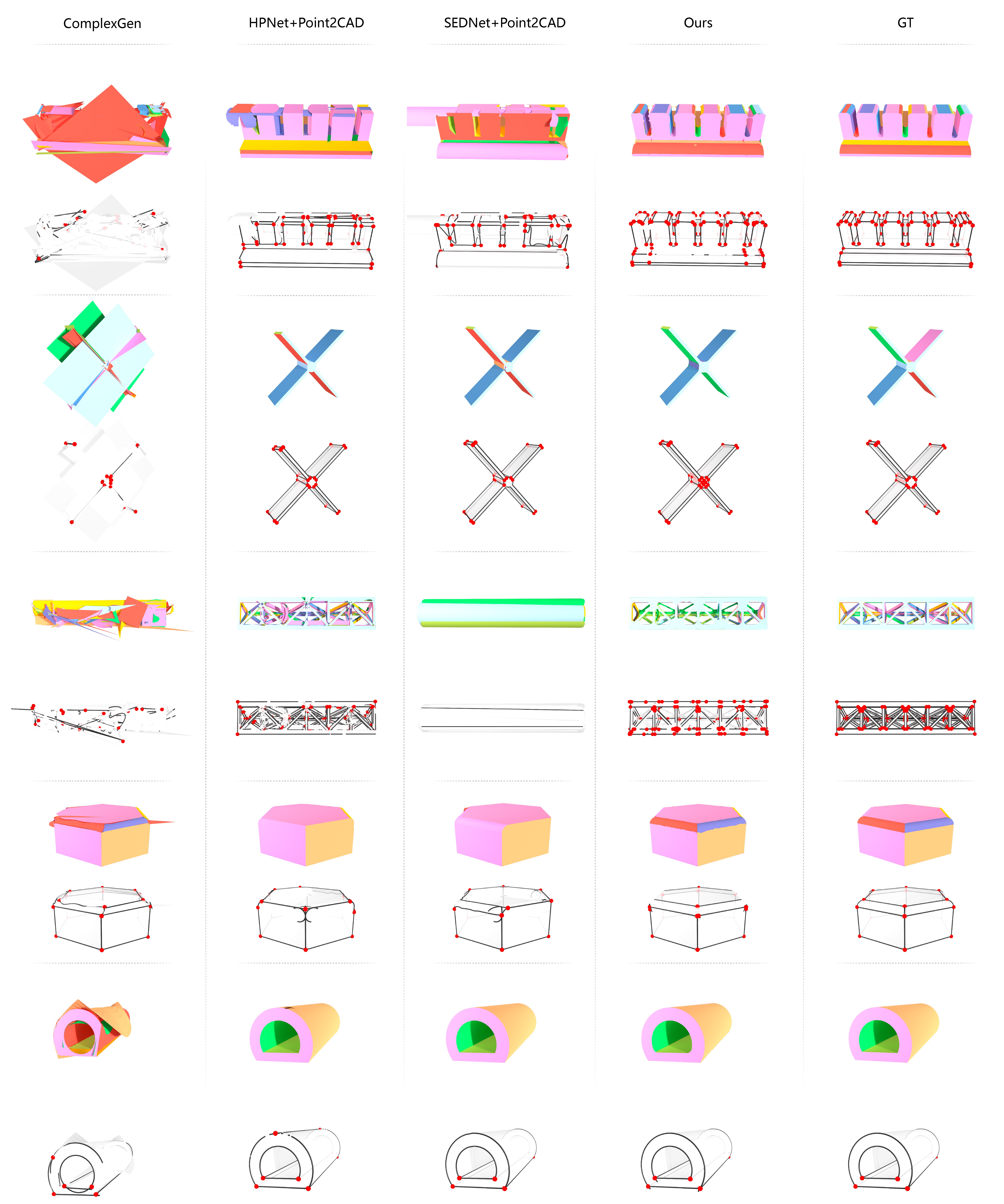}
    \caption{
        Qualitative comparisons of \textbf{representative} cases (5/6).
    }
    \label{fig:vis31_5}
\end{figure*}

\begin{figure*}
    \centering
    \includegraphics[width=0.95\linewidth]{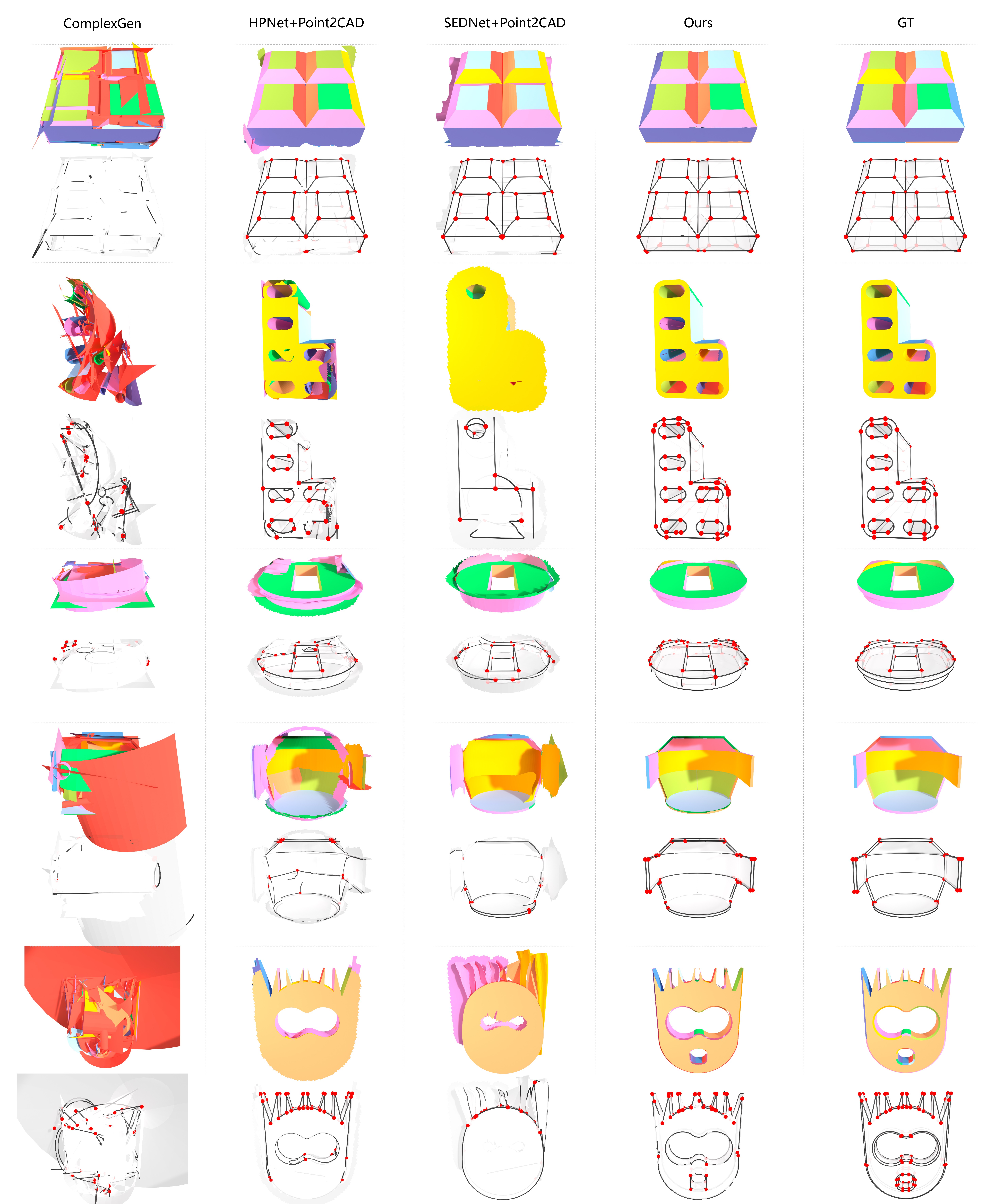}
    \caption{
        Qualitative comparisons of \textbf{representative} cases (6/6).
    }
    \label{fig:vis31_6}
\end{figure*}

\begin{figure*}
    \centering
    \includegraphics[width=0.95\linewidth]{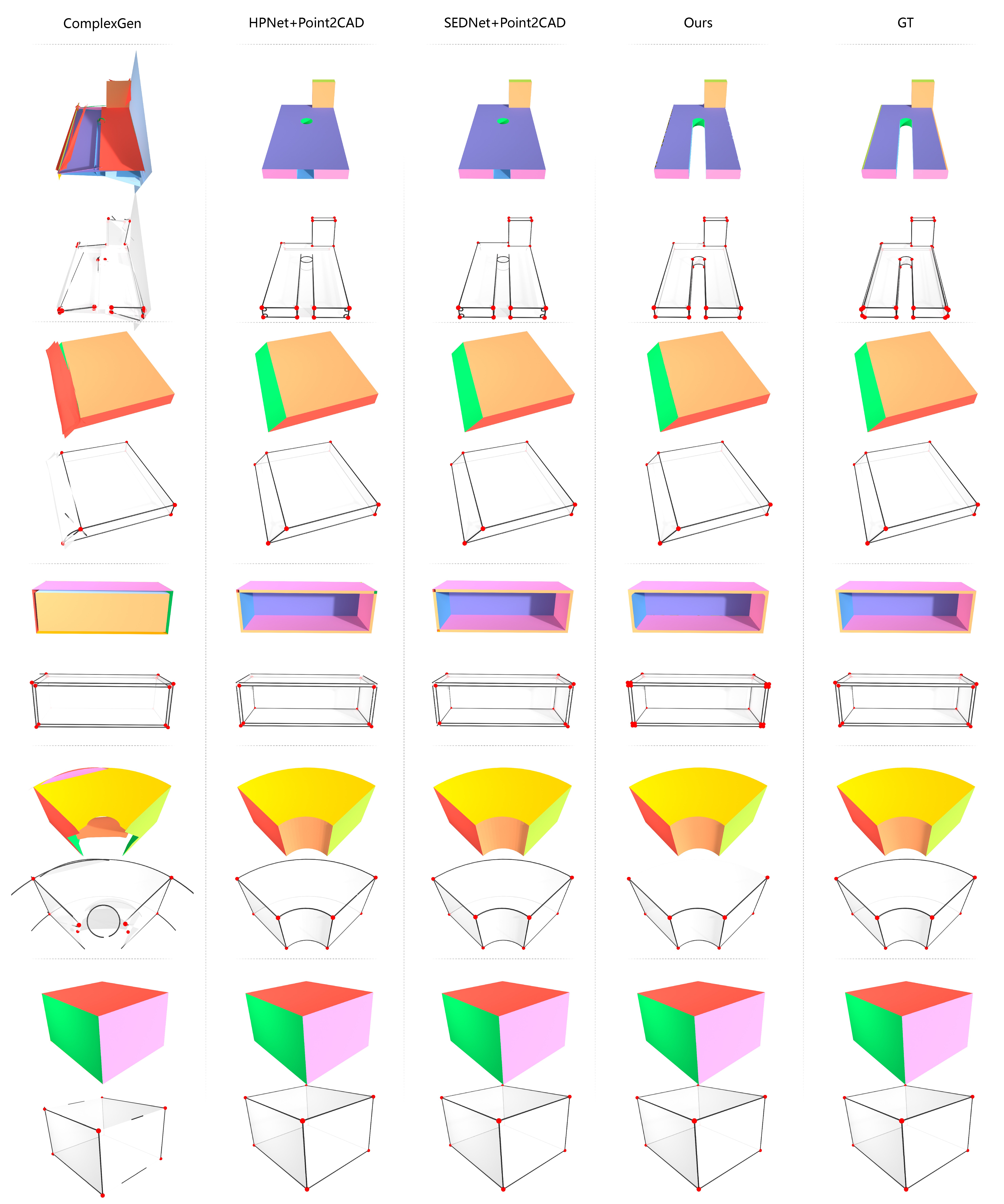}
    \caption{
        Qualitative comparisons of \textbf{randomly} selected cases (1/5).
    }
    \label{fig:vis_random_1}
\end{figure*}

\begin{figure*}
    \centering
    \includegraphics[width=0.95\linewidth]{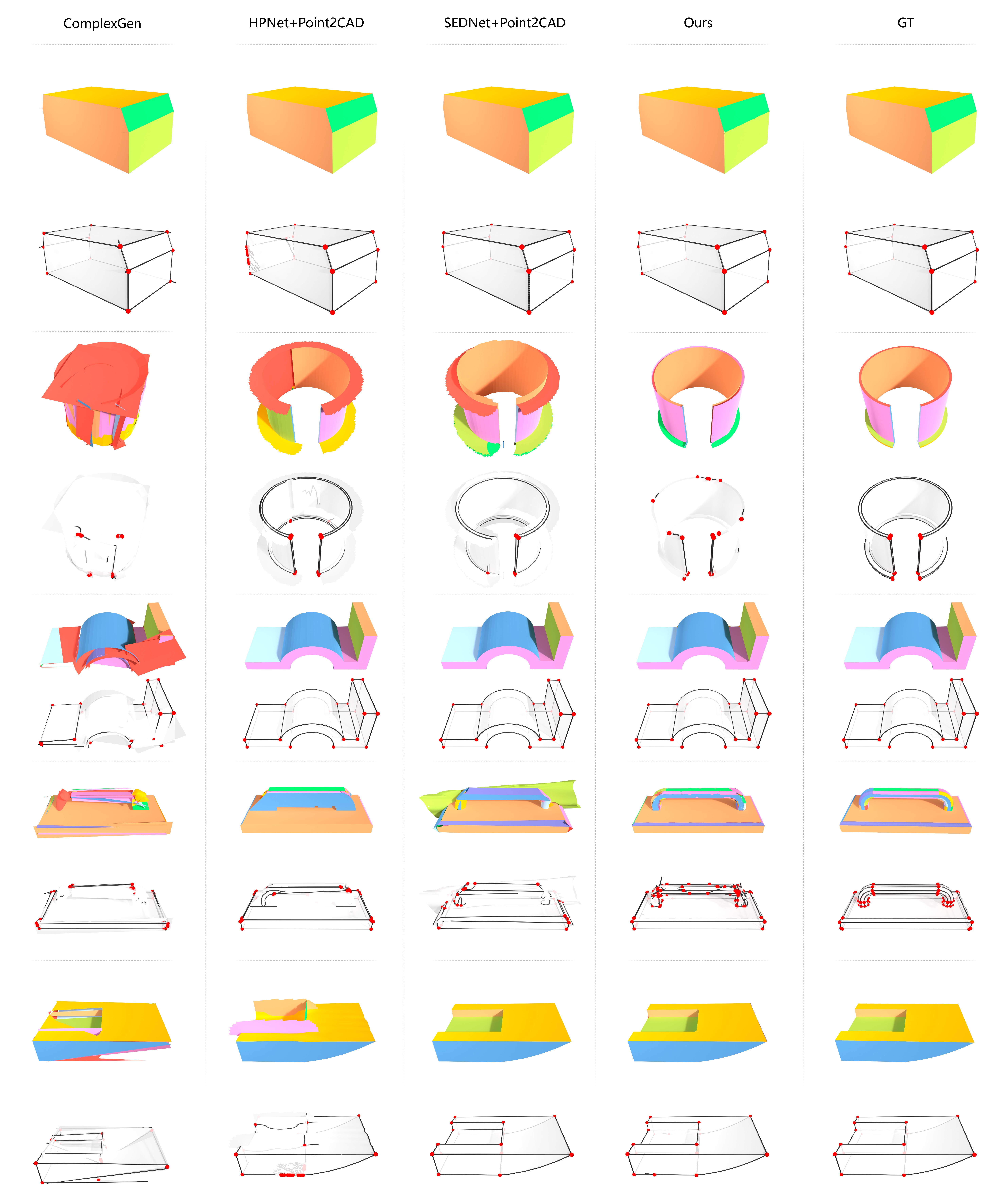}
    \caption{
        Qualitative comparisons of \textbf{randomly} selected cases (2/5).
    }
    \label{fig:vis_random_2}
\end{figure*}

\begin{figure*}
    \centering
    \includegraphics[width=0.95\linewidth]{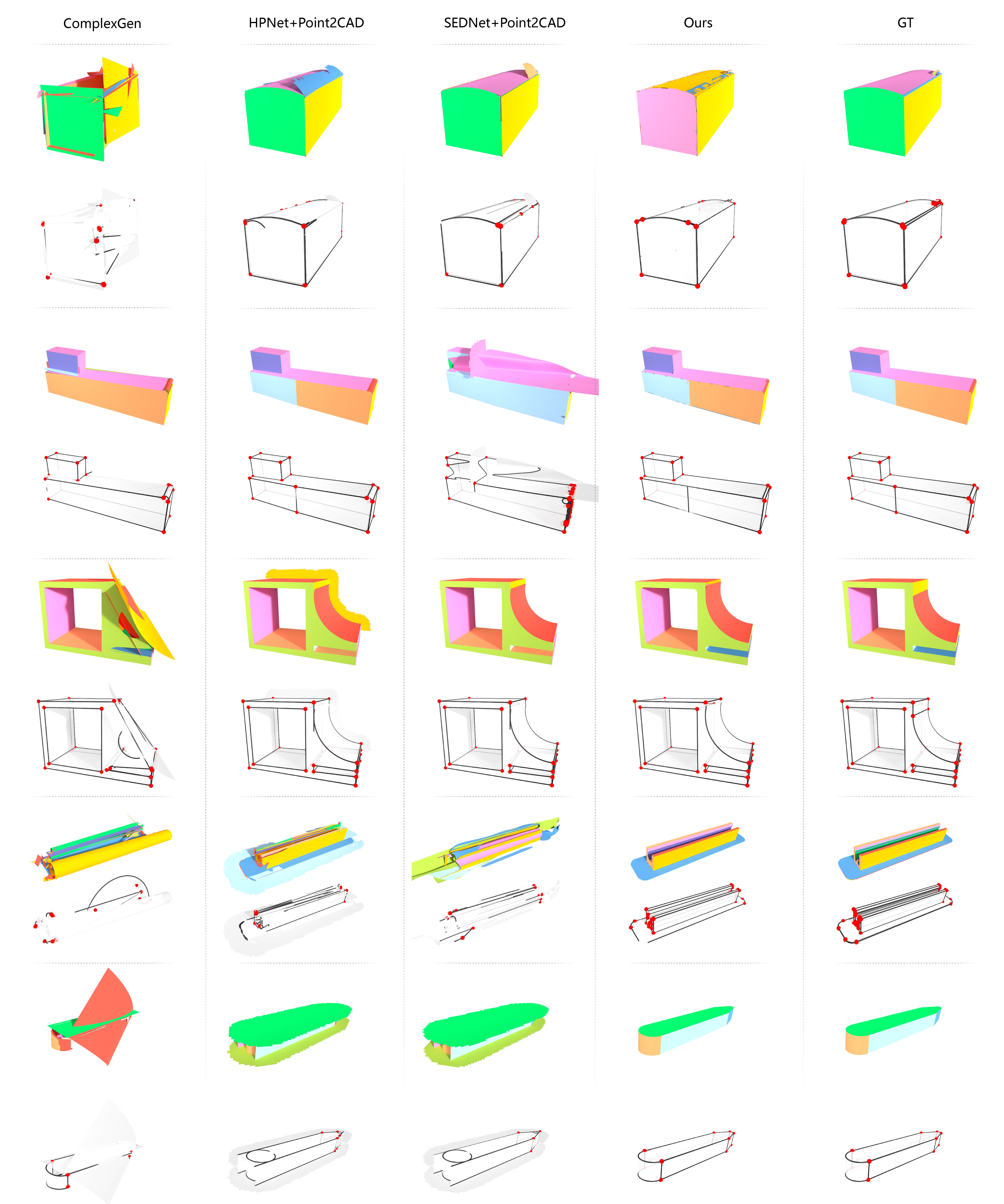}
    \caption{
        Qualitative comparisons of \textbf{randomly} selected cases (3/5).
    }
    \label{fig:vis_random_3}
\end{figure*}

\begin{figure*}
    \centering
    \includegraphics[width=0.95\linewidth]{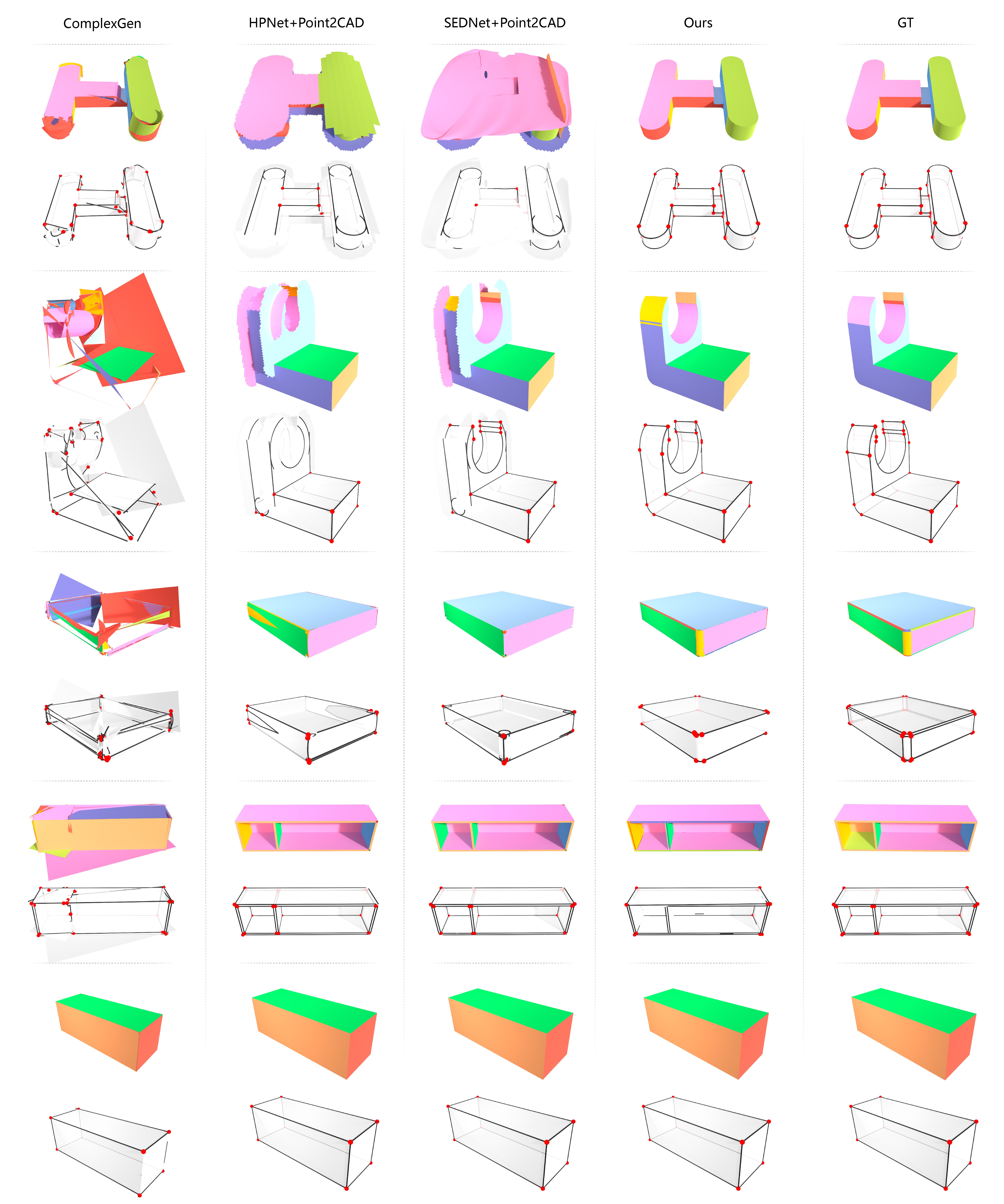}
    \caption{
        Qualitative comparisons of \textbf{randomly} selected cases (4/5).
    }
    \label{fig:vis_random_4}
\end{figure*}

\begin{figure*}
    \centering
    \includegraphics[width=0.95\linewidth]{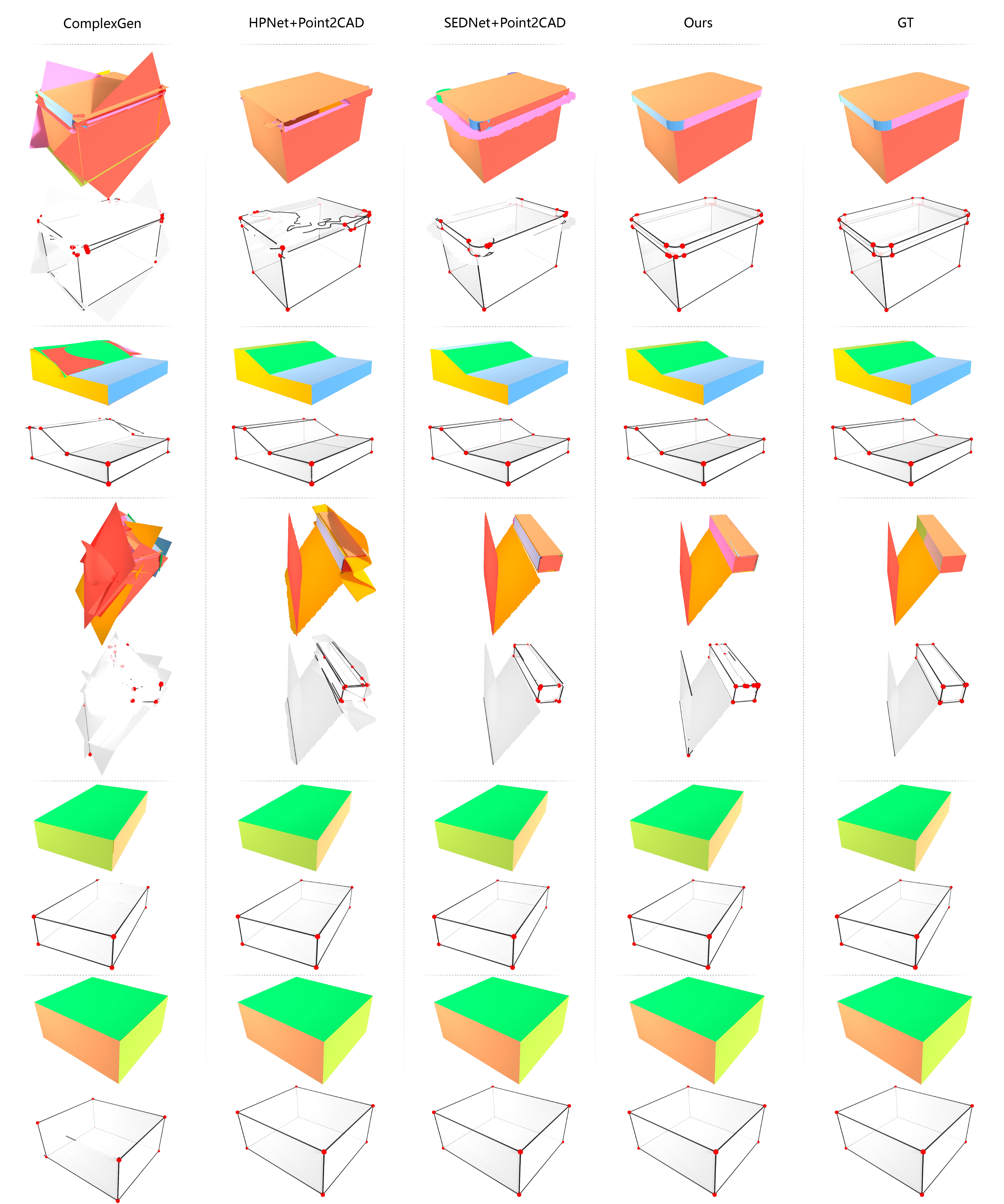}
    \caption{
        Qualitative comparisons of \textbf{randomly} selected cases (5/5).
    }
    \label{fig:vis_random_5}
\end{figure*}

We show the rest of \textbf{representative} shapes in
Figs.~\ref{fig:vis31_1},~\ref{fig:vis31_2},~\ref{fig:vis31_3},~\ref{fig:vis31_4},~\ref{fig:vis31_5},~\ref{fig:vis31_6}
and \textbf{randomly} selected shapes in
Figs.~\ref{fig:vis_random_1},~\ref{fig:vis_random_2},~\ref{fig:vis_random_3},~\ref{fig:vis_random_4},~\ref{fig:vis_random_5}.

\end{document}